\documentclass[lettersize,journal]{IEEEtran}
\usepackage{amsmath,amsfonts}
\usepackage{algorithmic}
\usepackage{algorithm}
\usepackage{array}
\usepackage[caption=false,font=normalsize,labelfont=sf,textfont=sf]{subfig}
\usepackage{booktabs}
\usepackage{textcomp}
\usepackage{stfloats}
\usepackage{url}
\usepackage{verbatim}
\usepackage{graphicx}
\usepackage{cite}
\usepackage{amssymb}
\usepackage{multirow}
\usepackage{booktabs}
\usepackage{graphicx}
\usepackage{color,xcolor}
\usepackage{hyperref}
\usepackage{threeparttable}
\usepackage{multirow}
\usepackage{amsmath}
\usepackage{CJKutf8}
\usepackage{color}
\makeatletter
\renewcommand{\maketag@@@}[1]{\hbox{\m@th\normalsize\normalfont#1}}%
\makeatother

\begin{document}
\begin{CJK}{UTF8}{gbsn}
\title{M2Restore: Mixture-of-Experts-based Mamba-CNN Fusion Framework for All-in-One Image Restoration}

\author{Yongzhen Wang, Yongjun Li, Zhuoran Zheng, Xiao-Ping Zhang, \textit{Fellow}, \textit{IEEE}, and Mingqiang Wei, \textit{Senior Member}, \textit{IEEE}

\thanks{Yongzhen Wang and Yongjun Li are with the School of Computer Science and Technology, Anhui University of Technology, Ma’anshan 243032, China (e-mail: wangyz@ahut.edu.cn; yongjun.li083@gmail.com).}
\thanks{Zhuoran Zheng is with the School of Cyberspace Security, Sun Yat-sen University, Shenzhen 518000, China (e-mail: zhengzr@njust.edu.cn).}
\thanks{Xiao-Ping Zhang is with the Tsinghua Shenzhen International Graduate School, Tsinghua University, Shenzhen 518055, China (e-mail: xpzhang@ieee.org).}
\thanks{Mingqiang Wei is with the School of Computer Science and Technology, Nanjing University of Aeronautics and Astronautics, Nanjing 210016, China, and also with the College of Artificial Intelligence, Taiyuan University of Technology, Taiyuan 030024, China (e-mail: mingqiang.wei@gmail.com).}




}

\markboth{Journal of \LaTeX\ Class Files,~Vol.~14, No.~8, August~2021}%
{Shell \MakeLowercase{\textit{et al}}: A Sample Article Using IEEEtran.cls for IEEE Journals}


\maketitle

\begin{abstract}
Natural images are often degraded by complex, composite degradations such as rain, snow, and haze, which adversely impact downstream vision applications. 
While existing image restoration efforts have achieved notable success, they are still hindered by two critical challenges: limited generalization across dynamically varying degradation scenarios and a suboptimal balance between preserving local details and modeling global dependencies.
To overcome these challenges, we propose M2Restore, a novel Mixture-of-Experts (MoE)-based Mamba-CNN fusion framework for efficient and robust all-in-one image restoration. M2Restore introduces three key contributions:
First, to boost the model’s generalization across diverse degradation conditions, we exploit a CLIP-guided MoE gating mechanism that fuses task-conditioned prompts with CLIP-derived semantic priors. This mechanism is further refined via cross-modal feature calibration, which enables precise expert selection for various degradation types.
Second, to jointly capture global contextual dependencies and fine-grained local details, we design a dual-stream architecture that integrates the localized representational strength of CNNs with the long-range modeling efficiency of Mamba. This integration enables collaborative optimization of global semantic relationships and local structural fidelity, preserving global coherence while enhancing detail restoration.
Third, we introduce an edge-aware dynamic gating mechanism that adaptively balances global modeling and local enhancement by reallocating computational attention to degradation-sensitive regions. This targeted focus leads to more efficient and precise restoration.
Extensive experiments across multiple image restoration benchmarks validate the superiority of M2Restore in both visual quality and quantitative performance. 
\end{abstract}


\section{Introduction}
\IEEEPARstart{I}{mage} restoration aims to reconstruct high-quality visual content from degraded observations, which is critical in numerous applications such as autonomous driving, traffic monitoring, and outdoor surveillance. Early efforts primarily relied on hand-crafted priors \cite{MengWDXP13}, \cite{TimofteDG13}, \cite{ZhuMS15}, \cite{KimK10},and physically grounded models \cite{He0T11} to guide the restoration process. While these approaches have demonstrated notable success in specific scenarios, their performance degrades considerably in real-world conditions involving diverse and complex degradations.
In recent years, deep learning-based approaches have attained significant advancements in mitigating weather-induced degradations such as rain streaks, haze, and snow \cite{ChenMSINR2024}, \cite{wang2024ucl}, \cite{Wang2023SmartAssign}. Nevertheless, the majority of existing methods remain task-specific, necessitating distinct models tailored to individual degradation types (e.g., APANet \cite{Wang2025APANet} for deraining, InvDSNet \cite{Quan2023InvDSNet} for desnowing). 
While these specialized models exhibit impressive efficacy within their respective domains, they demonstrate significant limitations when confronted with unseen or composite degradation scenarios. Furthermore, the requirement to retrain dedicated models for each degradation type incurs considerable computational overhead, posing a significant challenge for deployment on resource-constrained platforms. This issue is exacerbated by the combinatorial complexity inherent in real-world scenarios, where multiple degradations often coexist (e.g., concurrent rain streaks and atmospheric haze), rendering exhaustive model training both impractical and computationally prohibitive.

\begin{figure}[t] 
    \centering
    \includegraphics[width=1.0\linewidth]{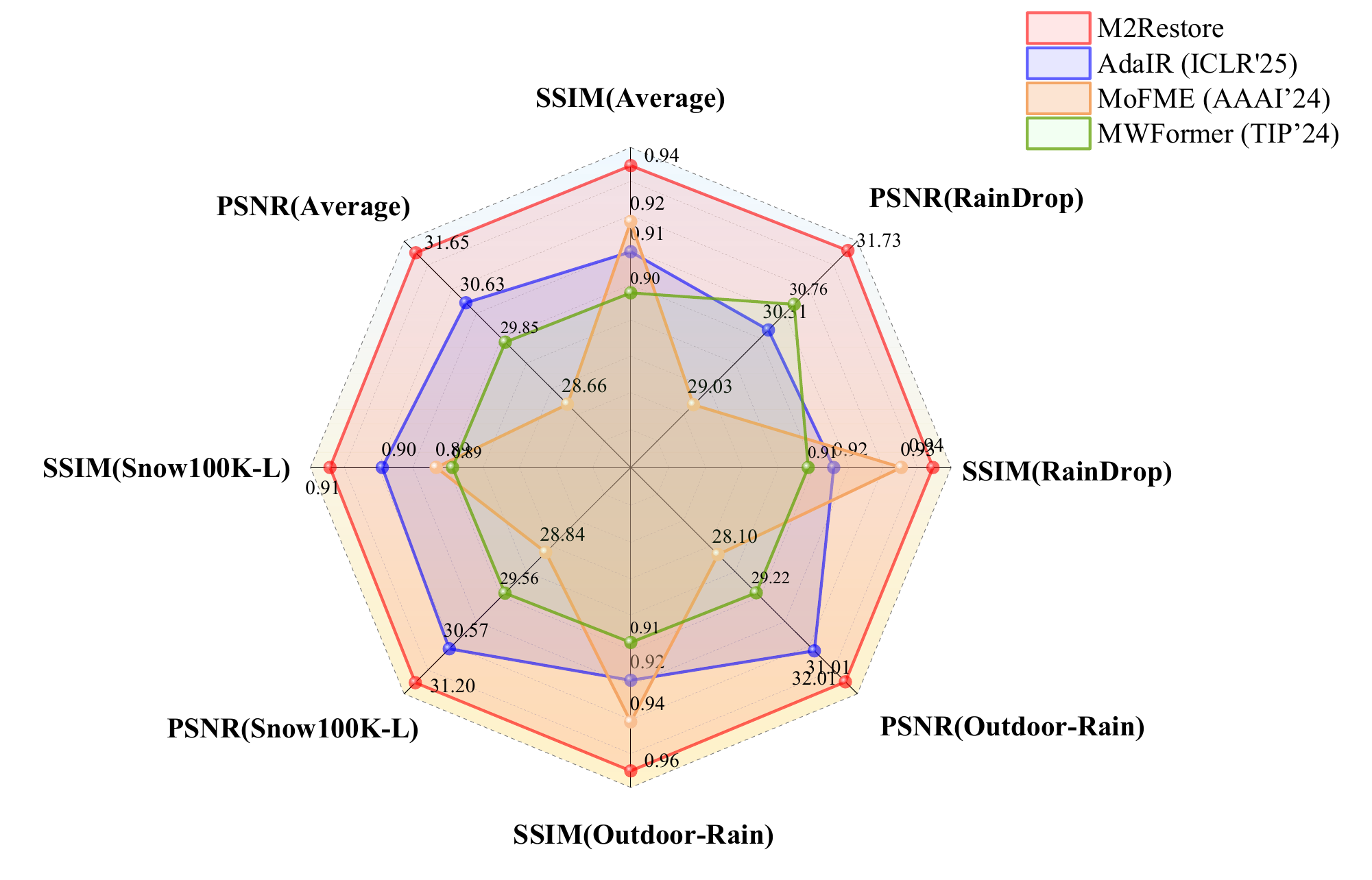}
	\caption{
    Quantitative Evaluation on the All-weather dataset \cite{LiTC20}: M2Restore ({\color{red} red}) vs. three state-of-the-art methods ({\color{blue}blue}, {\color[RGB]{244,166,102} yellow} and {\color{green}green}) under multi-degradation scenarios (rain/snow/average).}
 	\label{fig:fig1}
 \end{figure}

To overcome the limitation of single-degradation handling, a series of all-in-one image restoration frameworks \cite{Ai2024MPer-ceiver, Zhu2024MWDAT, ParkLC23, ZhangDMW25, 10453462, JiangZXG24, MWFormer} have recently been proposed, aiming to train unified models on mixed-degradation datasets. However, naive aggregation of multiple degradations often results in destructive interference during training, where overlapping artifacts generate complex distortion patterns that significantly impair restoration performance. Moreover, conventional CNN-based approaches, constrained by limited receptive fields, struggle to capture the long-range dependencies necessary for disentangling spatially intricate degradations. This limitation has driven the adoption of Transformer-based architectures \cite{LiangCSZGT21, ZamirA0HK022, cui2025adair}, which exhibit strong capabilities in modeling global contextual information and spatially adaptive restoration through self-attention mechanisms. Nonetheless, the quadratic complexity of self-attention necessitates patch-wise processing for high-resolution inputs, which disrupts feature continuity across patch boundaries. To alleviate these computational burdens, the emergence of State Space Models (SSMs), such as the Mamba architecture \cite{Gu2023Mamba}, offers an alternative with linear-time complexity (O(N)) for sequence modeling, enabling efficient long-range dependency capture in visual restoration tasks and inspiring works like MTAIR \cite{jiang2024multi}. However, Mamba's inherently sequential processing undermines spatial locality, resulting in gradual degradation of edge fidelity. Additionally, the parameter-sharing schemes intrinsic to both Transformer and Mamba architectures rely on static weight configurations, limiting their adaptability to diverse degradations and leading to considerable parameter redundancy and inference inefficiency, particularly at high resolutions.

The inherent limitations of the aforementioned architectural paradigms have spurred increased interest in Mixture-of-Experts (MoE) systems, which leverage sparse activation patterns and specialized subnetworks to effectively balance model capacity and computational efficiency. Originally introduced in the seminal work of \cite{JacobsJNH91}, the MoE framework has recently re-emerged as a promising solution for all-in-one image restoration, owing to its capacity to dynamically route inputs through expert modules via gating mechanisms. This design enables a single model to adaptively address diverse degradations such as raindrops, snow, and haze, within a unified architecture. Contemporary MoE implementations, such as Path-Restore \cite{YuWDTL22}, employ spatial gating networks to direct distinct degraded regions of an image to corresponding expert branches. However, the effectiveness of such systems is heavily contingent on the precision of the gating mechanism. Recent efforts have sought to enhance expert selection by integrating pre-trained CLIP models \cite{RadfordKHRGASAM21} to infuse textual priors into the routing process, as seen in DA2Diff \cite{xiong2025da2diffexploringdegradationawareadaptive}, LoRA-IR \cite{ai2024lora}, and DA-CLIP \cite{LuoG0SS24}. While this strategy improves degradation-type awareness, it often compromises content-aware feature extraction and fails to adequately preserve structural semantics (e.g., object boundaries and texture continuity), ultimately leading to perceptual distortions and degradation of content fidelity in the restored outputs.

In this paper, we introduce M2Restore, a novel Mixture-of-Experts-based Mamba-CNN fusion framework designed for all-in-one image restoration. Recognizing that Mamba's operation of flattening images into pixel sequences may disrupt the inherent spatial-local characteristics of images, we exploit a hybrid architecture termed Mamba-CNN Dual-Branch (MCDB). This design seamlessly integrates Mamba and CNN to facilitate dual-level interaction learning, capturing both patch-level and pixel-level representations. Additionally, to address the spatial heterogeneity of degradation-sensitive regions across different degradation types, we further incorporate a Dynamic Gated Feature Fusion (DGF) strategy into MCDB. This mechanism adaptively balances global context and local detail: Mamba pathways are emphasized in smooth regions for long-range dependency modeling, while CNN pathways are prioritized in edge-rich areas to preserve high-frequency structures. Moreover, inspired by the success of Mixture-of-Experts in multi-task learning, we introduce a Fine-Grained Degradation-Aware Dynamic Expert Routing (DDER) mechanism. This module leverages content-aware prompts alongside CLIP-derived semantic priors \cite{LuoG0SS24} to dynamically guide expert selection. Consequently, M2Restore is capable of performing precise and content-preserving restoration across a diverse range of degradation types. The comparative results presented in Fig. \ref{fig:fig1} underscores the superiority of M2Restore in addressing diverse degradations, consistently outperforming state-of-the-art methods.
In summary, the main contributions of this work are as follows:
\begin{itemize}

    \item We propose M2Restore, a novel unified MoE-based Mamba-CNN fusion framework for all-in-one image restoration. M2Restore flexibly reconstructs images affected by various degradation types without the need for retraining, leveraging dynamic expert selection.
    \item We introduce a dual-aware restoration mechanism by integrating trainable prompts with degradation priors from DA-CLIP, enabling M2Restore to jointly recognize degradation types and preserve semantic content for targeted and faithful reconstruction.
    \item We design a Mamba-CNN dual-branch architecture that combines the complementary strengths of Mamba and CNNs, supported by a dynamic gated fusion strategy. This architecture adaptively balances global context modeling and local detail refinement, resulting in high-fidelity image restoration.
\end{itemize}

\section{Related Work}
\subsection{Single Task Image Restoration}
Early restoration approaches are grounded in hand-crafted priors, such as non-local means \cite{BuadesCM05} and sparse coding \cite{olshausen1996emergence}, which exhibit limited generalization under complex degradation conditions. The advent of CNNs significantly improve performance on single-task restoration \cite{ZhangZCM017}, \cite{TianZWXZCL20}, yet their constrained receptive fields impede effective modeling of long-range degradation dependencies. Vision Transformers \cite{DosovitskiyB0WZ21} mitigate this limitation through self-attention mechanisms, enabling enhanced global context modeling. However, these gains come at the expense of substantial computational overhead (e.g., Restormer \cite{ZamirA0HK022} demands 367 GFLOPs for 256×256 images), posing a bottleneck for high-resolution image processing. Recently, state space models (SSMs), particularly Mamba \cite{Gu2023Mamba}, have emerged as promising alternatives, offering linear-complexity global modeling via selective state transitions. Although MambaIR \cite{GuoLDORX24} represents an early application of SSMs to image restoration, its reliance on naive image serialization strategies \cite{LiangCSZGT21}, which fragment images into fixed-length sequences, disrupts spatial coherence and introduces block artifacts along patch boundaries. 

\subsection{All-in-one Image Restoration}
Recently, all-in-one restoration frameworks have emerged as effective solutions for addressing composite and diverse degradations. Early attempts, such as Path-Restore \cite{YuWDTL22} leverage dynamic routing mechanisms, while AirNet \cite{LiLHW0022} introduces contrastive learning to facilitate degradation-aware feature extraction. The introduction of CLIP-driven paradigms marks a significant shift, beginning with \cite{LuoG0SS24}, which enables semantically-aware restoration through vision-language alignment. This direction is further expanded by PromptIR \cite{PotlapalliZ0K23} through learnable textual prompts. Despite their promise, these methods face two critical limitations: 1) text-alignment bias in prompt engineering, which limits robustness and generalization \cite{ArarVHAFPCS24}, and 2) computational inefficiency arising from the need for explicit degradation estimation pipelines. Meanwhile, transformer-based solutions like DeHamer \cite{guo2022dehamer} integrate degradation-aware modules but still require predefined degradation labels, restricting flexibility. Other dynamic architectures, such as conditional convolution networks \cite{ChenDLCYL20} introduce degradation-conditioned processing but often lack robust global dependency modeling.

\subsection{Mixture of Experts for Image Restoration}
\label{moe}
The Mixture of Experts (MoE) paradigm has evolved as a pivotal framework for enhancing model capacity via conditional computation. Early theoretical work by Jacobs et al. \cite{JacobsJNH91} introduce the concept of dynamically routing inputs to specialized subnetworks (“experts”), enabling efficient parameter scaling without proportional increases in computational overhead. In the context of image restoration, MoE has gained significant traction for its capacity to enable adaptive computation via expert subnetworks \cite{RenLLWGZZC24, abs-2503-15868, abs-2412-20157}. A notable contribution in this direction is Path-Restore \cite{YuWDTL22}, which pioneer content-aware patch routing with difficulty-aware rewards, dynamically assigning experts based on localized degradation complexity. More recent all-in-one restoration models \cite{zhang2025, zamfir2025, ai2024lora} further exploit MoE architectures to address diverse degradation types, using sparse gating mechanisms to coordinate CNN or Transformer experts for targeted artifact removal.

Despite their promise, existing MoE-based restoration approaches still face notable challenges. First, static expert initialization often biases the model towards specific training distributions, impairing its ability to generalize to unseen or mixed degradations. Second, conventional gating mechanisms \cite{kim2020restorings} tend to promote expert specialization, undermining inter-expert collaboration and resulting in discontinuous restoration artifacts. Although recent solutions such as contrastive expert regularization \cite{abs-2106-09667} and MoCE \cite{abs-2411-18466} have been proposed to mitigate these issues, they often come at the cost of increased complexity and additional hyperparameter tuning. 
 
\subsection{State Space Models for Image Restoration}
State Space Models (SSMs) have become a powerful framework for sequential data modeling, offering robust capabilities for capturing long-range dependencies through differential equations. Recent advancements have substantially expanded their computational efficiency and applicability \cite{TangWZ25, MatIR, abs-2404-11778, PengGS25}. Among these, the seminal Mamba architecture \cite{Gu2023Mamba} introduces selective state transitions to achieve linear-time complexity while preserving global receptive fields, marking a significant milestone in efficient long-range sequence modeling. Extending SSMs to the visual domain, Vision Mamba \cite{ZhuL0W0W24} processes 2D spatial structures as 1D sequences, enabling global modeling but sacrificing local structural fidelity. Hybrid approaches such as U-Mamba \cite{abs-2401-04722} attempts to mitigate this by cascading CNN and SSM modules in series. However, these models often lack effective cross-scale interaction due to their unidirectional information flow. More recent architectures have made notable progress. MaIR \cite{abs-2412-20066} incorporates a Sigmoidal scanning strategy and bidirectional cross-scanning paths to reduce redundancy and memory overhead during multi-directional processing, thereby improving restoration efficiency. Concurrently, MambaIRv2 \cite{abs-2411-15269} breaks the causal scanning constraint of traditional Mamba by integrating attention mechanisms, which significantly enhance long-distance pixel interactions and yield strong performance in super-resolution tasks. Despite these advancements, current SSM-based restoration methods still struggle to balance fine-grained spatial detail preservation with robust modeling of global degradations, especially under complex, composite degradation scenarios. 

\begin{figure*}[htbp] 
    \centering
    \includegraphics[width=0.95\linewidth]{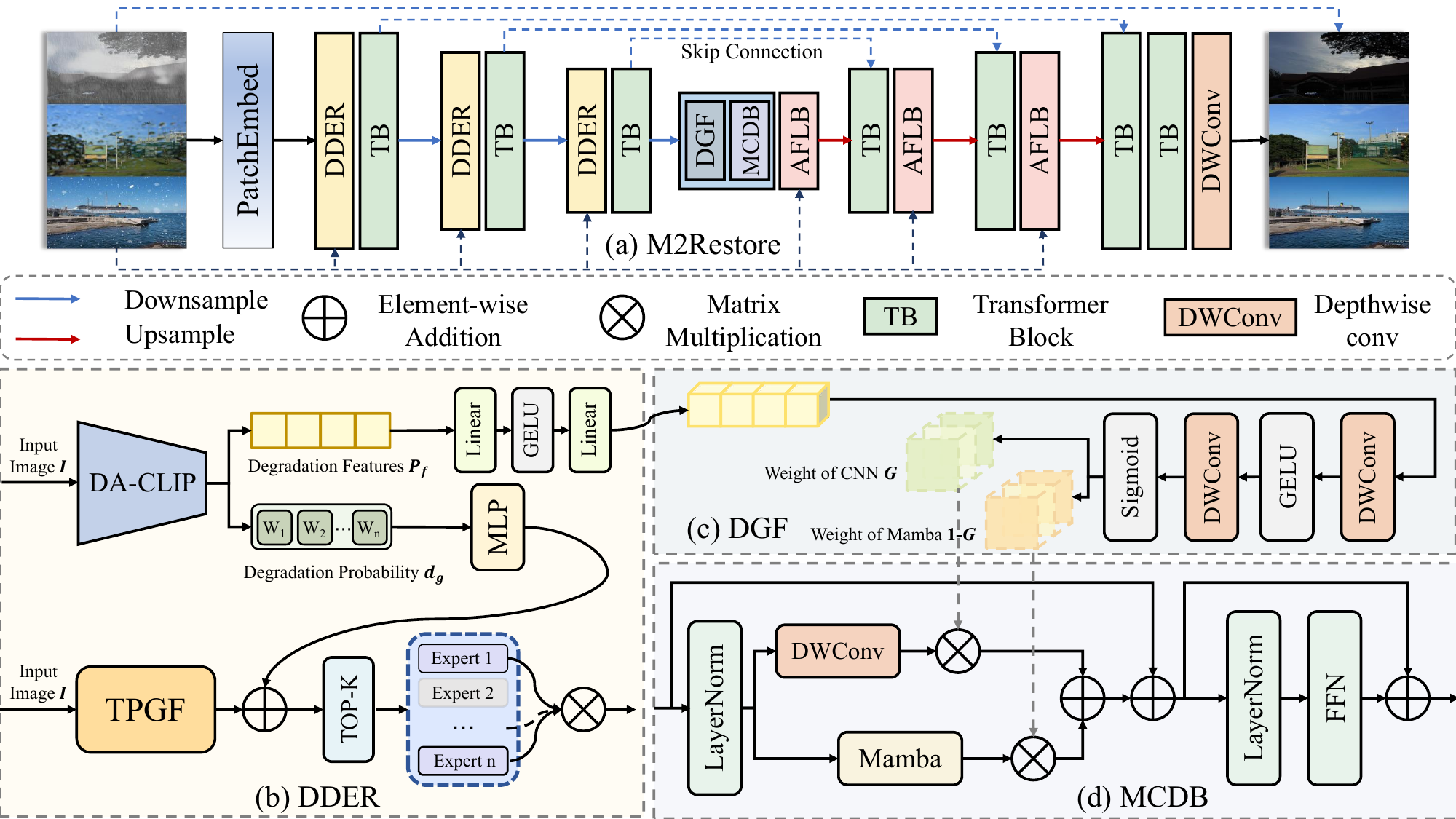}
	\caption{
The proposed M2Restore (a) is composed of three key components (b-d) designed to address the challenging all-in-one image restoration task: (b) Degradation-Aware Dynamic Expert Routing (DDER), responsible for selecting specialized processing experts based on the type of degradation; (c) Dynamic Gated Feature Fusion (DGF), embedded within the architecture to adaptively recalibrate feature responses and enhance focus on degradation-sensitive regions; and (d) Mamba-CNN Dual-Branch Hybrid Architecture (MCDB), which synergistically integrates CNN's local feature extraction with Mamba's global modeling capabilities to achieve high-precision restoration. AFLB denotes the
Adaptive Frequency Learning Block.}
 	\label{fig:fig2}
 \end{figure*}

\section{Method}   
In this section, we present a comprehensive description of the proposed M2Restore framework, specifically designed to tackle diverse weather-induced degradations. The overall architecture of M2Restore is depicted in Fig. \ref{fig:fig2}.

\subsection{Overview}
The proposed M2Restore framework concurrently tackles the challenges of multi-degradation generalization and high-frequency detail preservation through a hybrid encoder-decoder architecture. As depicted in Fig. \ref{fig:fig2}, given a degraded image, the model begins by applying a convolutional patch embedding layer to extract low-level feature representations, followed by a hierarchical computational pipeline. The shallow encoder layers employ Transformer blocks with gradually increasing depth to ensure model performance, while the deepest layer integrates Mamba for global correlation modeling with linear complexity, complemented by CNN-based local feature enhancement via a Dynamic Gated Fusion (DGF) mechanism. This integration effectively harmonizes global degradation modeling with the preservation of fine-grained local structures. At the core of the framework lies the Degradation-Aware Dynamic Expert Routing (DDER) module, which adaptively assigns restoration experts through task-aware prompts and CLIP-derived degradation priors. Moreover, to further enrich the restored image details, we incorporate frequency domain information into the spatial domain via the Adaptive Frequency Learning Blocks (AFLB) \cite{cui2025adair}, thereby facilitating refined detail restoration. The following sections provide an in-depth exposition of key components, including the expert assignment mechanism in DDER and the fusion strategy governed by DGF.

\subsection{Dynamic Degradation-Aware Expert Router}
As discussed in Section \ref{moe}, Mixture-of-Experts (MoE)-based architectures have demonstrated superior performance over conventional baseline models. To this end, we design the Dynamic Degradation-aware Expert Routing module (DDER) module, whose primary objective is to endow the network with two key capabilities via task-driven adaptive computation: (1) high-precision restoration of known degradation typrs; and (2) strong generalization to unseen degradations. As illustrated in Fig. \ref{fig:fig2}(b), DDER is composed of three co-optimized stages:

\subsubsection{Stage One: Multimodal Prompt Generation}
\begin{figure}[t] 
    \centering
    \includegraphics[width=1.0\linewidth]{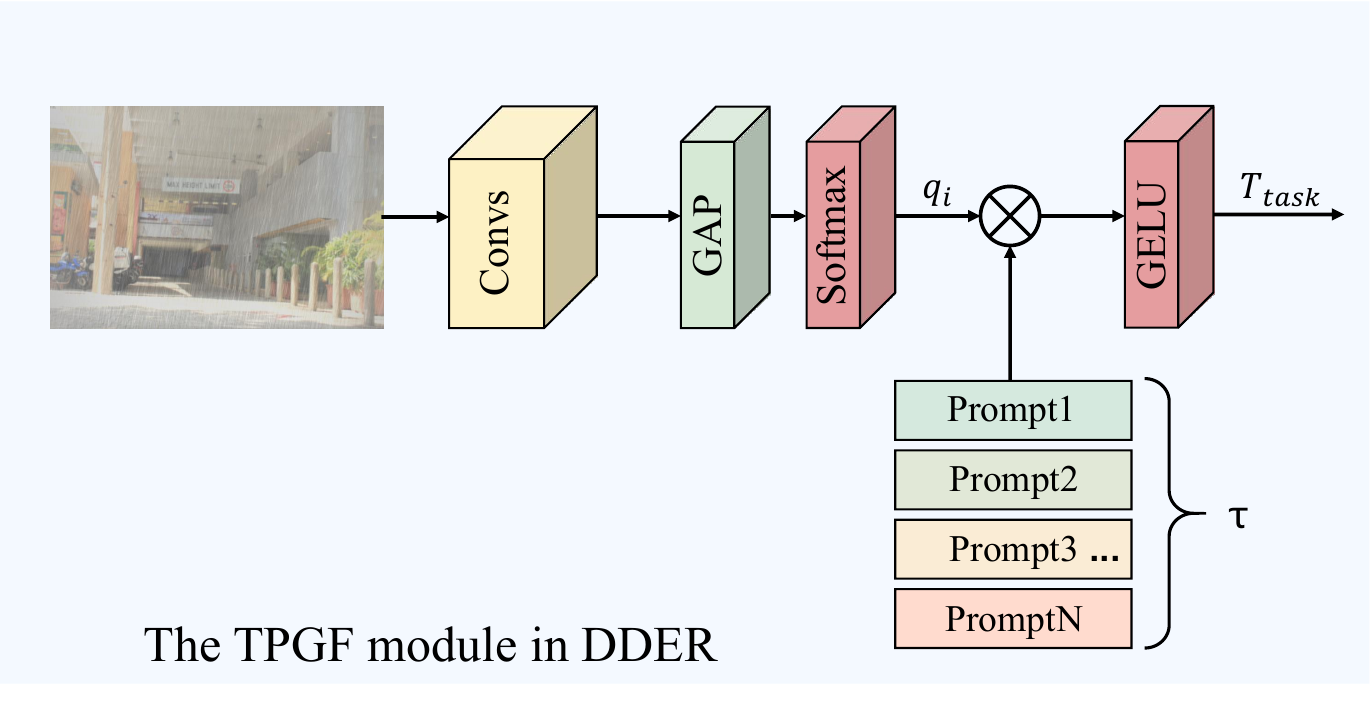}
	\caption{
	Overall architecture of the prompt-driven framework.}
 	\label{fig:fig3}
 \end{figure}

This stage enables synergistic integration of task specificity and degradation generalization through a dual-stream architecture: 

\textbf{Task Prompt Generation Flow}: Given the input feature map $\textit{\textbf{I}} \in \mathbb{R}^{H \times W \times 3}$, where $d$ denotes the channel dimensionality, the features are progressively compressed through a cascade of $3\times3$ convolutions. The resulting representation is then projected into a $\mathbb{R}^{C}$-dimensional latent space via global average pooling followed by a linear transformation, aligning it with the dimensionality of the base prompt library. A Softmax function is subsequently applied to obtain the attention weights $\textit{\textbf{q}}_\textit{i} \in \mathbb{R}^{H \times W \times C}$, which reflect the relevance of each base prompt to the current task. These weights are employed to retrieve task-relevant prompt embeddings $\textit{\textbf{T}}_{\textit{task}} \in \mathbb{R}^{H \times W \times M}$ from the predefined base prompt library $\tau = \{\mathbf{t}_1, ..., \mathbf{t}_C\}^T \in \mathbb{R}^{C \times M}$. Finally, $\textit{GELU}$ activation is employed to introduce nonlinearity and enrich representational expressiveness. As illustrated in Fig. \ref{fig:fig3}, the entire process can be expressed as:
\begin{equation}
\textit{\textbf{q}}_\textit{i} = \textit{softmax}\Big( \textit{\textbf{W}}_\textit{l} \cdot {\textit{GAP}\big( \textit{Convs}(\textit{\textbf{I}}) \big)},
\end{equation}
\begin{equation}
\textit{\textbf{T}}_{\textit{task}} = \textit{GELU}( \textit{\textbf{q}}_\textit{i}\cdot \tau).
\end{equation}
where $\text{Convs}$ denotes a sequence of convolutional operations, $\textit{\textbf{W}}_\textit{l} \in \mathbb{R}^{C^{'} \times C}$ represents a learnable linear projection matrix, $\tau \in \mathbb{R}^{C \times M}$ refers to the learnable base prompt library, and $\textit{GAP}$ indicates the global average pooling operation.

\textbf{Degradation Prior Extraction Stream}: To mitigate the overfitting risk and limited generalization inherent in heavily trained prompts, and to enhance sensitivity to previously unseen degradations, we deviate from the MEASNet \cite{MEAS} paradigm by incorporating a frozen DA-CLIP \cite{LuoG0SS24} model to process the original degraded image $\textit{\textbf{I}} \in \mathbb{R}^{ H \times W \times3}$. This integration equips the framework with the capacity to discern both known and unknown degradation types. As depicted in Fig. \ref{fig:fig2}(b), we illustrate examples of degradation-type probability distributions generated from various degraded inputs. Specifically, the degraded image $\textit{\textbf{I}}$ is processed by the frozen DA-CLIP model \cite{LuoG0SS24}, producing two outputs: a global degradation probability vector $\textit{\textbf{d}}_\textit{g} =\left[ w_{1}, w_{2}...w_{n}  \right]\in\mathbb{R}^D$, where each $w_n$ denotes the likelihood associated with a specific known or unknown degradation type, and a spatial degradation feature vector $\textit{\textbf{P}}_\textit{f}\in \mathbb{R}^{1 \times 512}$. The global degradation probability $\textit{\textbf{d}}_\textit{g}$ is subsequently utilized to guide the adaptive restoration of $\textit{\textbf{I}}$ within our framework. 

\subsubsection{Stage Two: Bias-Enhanced Expert Routing}

To effectively integrate the task prompt $\textit{\textbf{T}}_{\textit{task}}$ from Stage One with the degradation prior $\textit{\textbf{d}}_\textit{g}$, we introduce a dual-modal joint routing mechanism. Specifically, to ensure that the routing network simultaneously captures fine-grained spatial features and global task semantics, the task prompt $\textit{\textbf{T}}_{\textit{task}} \in \mathbb{R}^{H \times W \times M}$ is first projected to match the dimensionality of the input feature $\textit{\textbf{x}} \in \mathbb{R}^{H \times W \times d}$, yielding $\textit{\textbf{T}}_{\textit{task}} \in \mathbb{R}^{H \times W \times d}$. These two feature maps are then concatenated along the channel axis to construct a unified joint representation: 
\begin{equation}
\textit{\textbf{x}}_\textit{p} = [\textit{\textbf{x}}; \textit{\textbf{T}}_{\textit{task}}] \in \mathbb{R}^{ H\times W\times 2d}.
\end{equation}

Subsequently, for computational efficiency, the spatial dimensions $H$ and $W$ dimensions into $P = H\times W$. The base routing score and degradation-aware bias are then computed in parallel, facilitating efficient expert selection while preserving spatially contextualized information:
\begin{equation}
\textit{\textbf{score}} = \textit{\textbf{x}}_\textit{p} \textit{\textbf{W}}_\textit{g} \quad \scriptstyle{(\textit{\textbf{W}}_g \in \mathbb{R}^{2d \times N})},
\end{equation}
\begin{equation}
\textit{\textbf{b}} = \textit{\textbf{d}}_\textit{g} \textit{\textbf{W}}_b \quad \scriptstyle{( \textit{\textbf{W}}_b \in \mathbb{R}^{D \times N})},
\end{equation}
where $\textit{\textbf{score}} \in \mathbb{R}^{P \times N}$ denotes the base routing scores, $\mathbf{b} \in \mathbb{R}^{P \times N}$ represents the degradation-aware bias, $P$ is the total number of spatial positions (i.e., pixels), and $N$ is the number of available experts. The vector $\mathbf{\textit{d}}_\textit{g} \in \mathbb{R}^D$ encodes the global degradation probabilities. The final dynamic fusion is modulated by a learnable gating coefficient $\alpha$, enabling adaptive and content-aware expert selection conditioned on both local features and global degradation context:
\begin{equation}
\textit{\textbf{S}} = \textit{\textbf{score}} + \alpha \odot \textit{\textbf{b}}. \quad \alpha \in \mathbb{R}^N .
\end{equation}

An input-dependent noise generation module is introduced to enhance the robustness and diversity of routing decisions. Specifically, the joint feature $\textit{\textbf{x}}_\textit{p}$ is first projected into the noise parameter space, and the resulting noise intensity is constrained to be non-negative by applying the Softplus activation function: 
\begin{equation}
\sigma = \textit{Softplus}(\textit{\textbf{x}}_\textit{p} \textit{\textbf{W}}_\textit{n}) \in \mathbb{R}^K, \quad \textit{\textbf{W}}_\textit{n} \in \mathbb{R}^{2d \times N},
\end{equation}
where $\textit{\textbf{W}}_\textit{n}$ is a learnable parameter used to adaptively generate noise intensity from the joint feature $\textit{\textbf{x}}_\textit{p}$. Moreover, to enable adaptive perturbation to the routing scores during training and enhance routing performance, a random mask vector is introduced via Bernoulli sampling:
\begin{equation}
\epsilon \sim \textit{Bernoulli}(p=0.5), \quad \epsilon \in \{0,1\}^N.
\end{equation}

Subsequently, Gaussian noise injection is applied to the selected dimensions, introducing controlled perturbations to the routing scores:
\begin{equation}
\tilde{\textit{\textbf{S}}} = \textit{\textbf{S}} + \epsilon \odot \mathcal{N}(0, \sigma^2).
\end{equation}

This strategy facilitates the exploration of potentially superior expert combinations during training, while preserving the differentiability required for gradient backpropagation. During inference, the noise is deactivated ($\epsilon=0$), ensuring deterministic expert selection and stable output generation.

\subsubsection{Stage Three: Sparse Expert Activation}

Considering that different pixels are affected by degradation in varying degrees, we implement pixel-wise adaptive restoration based on the noise-perturbed expert scores generated in Stage Two. We define a set of $N$ candidate experts ${\{\textit{\textbf{F}}_\textit{n}|\textit{n} = 1,..., N\}}$, where each expert ${F_n}$ is specialized in reconstructing features associated with a particular type of degradation. The perturbed expert scores $\tilde{\mathbf{S}}\in\mathbb{R}^{P \times N}$ are refined using a Top-K selection strategy along the expert dimension. Specifically, for each pixel $(i,j)$, the Top-K function identifies the $K$ most suitable experts based on their routing scores. These selected scores are then normalized via a Softmax function to obtain sparse, differentiable expert weights. To reduce computational overhead, only the top $K$ experts are activated per pixel, while the remaining weights are zeroed out via the \textit{Top$K(\cdot)$} operation. The final per-pixel normalized selection scores ${Se\in\mathbb{R}^{P \times N}}$ are computed as:
\begin{equation}
\textit{\textbf{Se}} = \textit{softmax}(\textit{Top}K(\tilde{\mathbf{S}},K)),
\end{equation}
\begin{equation}
\textit{Top}K(v,K) = \begin{cases}
v, & \text{if } v \text{ is in the top } K \text{ elements} \\
0, & \text{otherwise}
\end{cases},
\end{equation}
where $\textit{\textbf{Se}}({i,j})$ represents the degree to which the pixel at position $i$ requires the the $j$-th expert for accurate restoration. The higher value of $\textit{\textbf{Se}}({i,j})$ indicates a stronger reliance on the corresponding expert. Rather than engaging all available experts, we adopt a sparse activation strategy by selecting the top-$K$ experts with the highest routing scores for each pixel. This selective expert activation enhances the restoration capability of the most relevant experts while suppressing interference from unrelated or weakly correlated tasks. Moreover, this mechanism naturally facilitates expert sharing across related image restoration tasks. Since different degradation types may activate overlapping subsets of experts, task affinity is implicitly encoded. Such shared expert utilization not only promotes parameter efficiency but also enables knowledge transfer between tasks, thereby improving the overall performance and generalization capability of the model.

\subsection{Mamba-CNN Dual-Branch Hybrid Architecture}
As illustrated in Fig. \ref {fig:fig2}(d), we proposes a Mamba-CNN dual-branch hybrid architecture, which efficiently captures multi-scale features by jointly modeling global degradation contexts and local structural details. Given an input feature map $\textit{\textbf{F}} \in \mathbb{R}^{H \times W \times C}$, we first apply \textit{LayerNorm} to normalize cross-channel features. The normalized features are then fed into the dual-branch feature extraction module. In the CNN branch (local detail pathway), we employ depthwise separable convolution to focus on extracting high-frequency texture details, thereby preserving fine-grained structural information:
\begin{equation}
\textit{\textbf{F}}_{\textit{cnn}} =  \textit{GELU}(\textit{DWConv}_{3\times3}(\textit{LayerNorm}(\textit{\textbf{F}}))).
\end{equation}

Among them, the depthwise convolution (\textit{DWConv}) in the CNN branch preserves fine-grained edge structures through channel-wise independent computation, yielding the output $\textit{\textbf{F}}_\textit{cnn}$. In the Mamba branch, which serves as the global dependency modeling, the Mamba module is leveraged for its powerful capability in handling long sequential data with linear computational complexity. This capability enables it to capture long-range pixel interactions that are prohibitively expensive to model using conventional self-attention mechanisms. Specifically, the input feature $\textit{\textbf{F}}$ is first normalized using \textit{LayerNorm} and then reshaped and transposed to the format $(B, C, L)$, where $L=H×W$ represents the flattened spatial dimensions. This sequence is then fed into the SSM layer to model global dependencies. Following this, another Layer Normalization is applied, and the output is finally reshaped back to the original spatial format $(B, C, H, W)$: 
\begin{equation}
\textit{\textbf{F}}' = \textit{Transpose}(\text{Reshape}(\textit{\textbf{F}}, (B, C, L))),
\end{equation}
\begin{equation}
\textit{\textbf{F}}'' =\textit{Norm} (\textit{SSM}(\textit{Norm}(\textit{\textbf{F}}'))),
\end{equation}
\begin{equation}
\textit{\textbf{F}}_\textit{mamba} =\textit{Reshape}(\textit{Transpose}(\textit{\textbf{F}}), (B, C, H, W)).
\end{equation}

This pathway equips the model with a robust mechanism for modeling broad spatial interactions, complementing the local detail refinement of the CNN branch.

Moreover, to adaptively balance fine-grained local detail with holistic global context, we introduce a novel Edge-Aware Dynamic Gating mechanism (DGF), as depicted in Fig. \ref{fig:fig2}(c). The spatial degradation feature $\textit{\textbf{P}}_\textit{f}$, extracted in the preceding stages, is first projected to match the dimensionality of the feature map $\textit{\textbf{F}}$. These two representations are then concatenated along the channel axis and passed through the DGF module to compute dynamic gating weights $G \in \mathbb{R}^K$. The operation is formally defined as: 
 \begin{equation}
G = \sigma\left( \textit{Conv}_{1\times1}\left( [\textit{\textbf{F}},  \textit{\textbf{P}}_\textit{f}] \right) \right),
\end{equation}
\begin{equation}
\textit{\textbf{F}}_{\textit{out}} = G \odot \textit{\textbf{F}}_{\textit{cnn}} + (1 - G) \odot \textit{\textbf{F}}_{\textit{mamba}}.
\end{equation}

Here, $G$ denotes the initial dynamic gating weight assigned to the CNN branch, while its complementary weight $(1-G)$ is allocated to the Mamba branch. The final output feature $\textit{\textbf{F}}_{\text{out}}$ is computed as a weighted combination of both branches.

\subsection{Loss Functions}
We optimize our network using a composite loss function comprising the L1 loss $\mathcal{L}_{\textit{L1}}$ and the coefficient of variation squared loss  $\mathcal{L}_{\textit{balance}}$. The L1 loss penalizes the absolute deviation between the restored image and the ground-truth clean image, thereby encouraging pixel-wise accuracy. It is formally defined as:
\begin{equation}
\mathcal{L}_{\textit{L1}} =\frac{1}{N} \sum_{i=1}^{N} \left| y_i - \hat{y}_i \right| ,
\end{equation}
where $N$ denotes the total number of pixels, $y_{i}$ is the ground-truth value of the $i_{th}$ pixel, and $\hat{y}_{i}$ is the corresponding restored prediction. The balance loss $\mathcal{L}_\textit{balance}$ is defined as:
\[ \mathcal{L}_{\textit{balance}} = \frac{\textit{Var}(\textit{\textbf{w}})}{(\textit{Mean}(\textit{\textbf{w}}))^2 + \epsilon} + \frac{\textit{Var}(\textit{\textbf{s}})}{(\textit{Mean}(\textit{\textbf{s}}))^2 + \epsilon} ,\]
where $\textit{\textbf{w}}=\left[ \sum_{j=1}^{n} w_{1j}, \sum_{j=1}^{n} w_{2j}, \ldots, \sum_{j=1}^{n} w_{N_{\text{experts}}j} \right]$ denotes the aggregated selection weight of each expert across all pixels, $n$ is the total number of pixels, $N_\text{experts}$ is the number of experts and $\textit{\textbf{s}}=\left[ s_{1}, s_{2}...s_{n}  \right]$ represents the sparse activation count, i.e., the number of times each expert is selected. $\textit{Var($\cdot$)}$ and $\textit{Mean($\cdot$)}$ denote the variance and mean operations, respectively, and $\epsilon$ is a small positive constant introduced to ensure numerical stability during division. Finally, The total $\mathcal{L}_\textit{total}$ is formulated as:
\begin{equation}
\mathcal{L}_{\textit{total}} =\mathcal{L}_{\textit{L}1} + \lambda\mathcal{L}_{\textit{balance}},
\end{equation}
where ${\lambda}$ is a hyperparameter that modulates the relative contribution of the balance loss within the training phase.

\begin{table*}[t]
\renewcommand\arraystretch{1.2}
	\centering
	\caption{Quantitative comparison on the All-weather dataset. The best and second-best values are highlighted in bold and underlined, respectively}
	\label{Tab:tab1}
	\begin{threeparttable}
		\footnotesize
		\centering
		\setlength{\tabcolsep}{2.0mm}{
			\begin{tabular}{ccccccccccc}
				\toprule
                    \multirow{2}{*}{Type}&
				\multirow{2}{*}{Method}&
				\multirow{2}{*}{Publication}&
				\multicolumn{2}{c}{Outdoor-Rain \cite{LiCT19}}&
                    \multicolumn{2}{c}{Snow100K-L \cite{LiuJHH18}}&
                    \multicolumn{2}{c}{RainDrop \cite{QianTYS018}}&
				\multicolumn{2}{c}{Average}\cr
				\cmidrule(lr){4-5} \cmidrule(lr){6-7} \cmidrule(lr){8-9} \cmidrule(lr){10-11}&
				  & & PSNR$\uparrow$ & SSIM$\uparrow$  & PSNR$\uparrow$ & SSIM$\uparrow$  & 
                    PSNR$\uparrow$ & SSIM$\uparrow$  & PSNR$\uparrow$ & SSIM$\uparrow$\cr
				\midrule
                    \multirow{6}{*}{\centering General} 
                    & MPRNet \cite{ZamirA0HK0021} & CVPR'21 & 28.08 & 0.931 & 27.92 & 0.911 & 29.45 & 0.942 & 28.48 & 0.928 \cr
                    & SwinIR \cite{LiangCSZGT21} & ICCVW'21 & 23.23 & 0.869 & 28.18 & 0.880 & 30.82 & 0.904 & 27.41 & 0.884 \cr
                    & NAFNet \cite{ChenCZS22} & ECCV'22 & 23.21 & 0.840 & 27.68 & 0.847 & 28.90 & 0.890 & 26.60 & 0.859 \cr
                    & Uformer \cite{WangCBZLL22} & CVPR'22 & 25.40 & 0.889 & 26.60 & 0.887 & 27.38 & 0.919 & 26.46 & 0.898  \cr
                    & Restormer \cite{ZamirA0HK022} & CVPR'22 & 27.24 & 0.921 & 27.76 & 0.907 & 29.29 & 0.937 & 28.10 & 0.922 \cr
                    & GRL \cite{LiFXDRTG23} & CVPR'23 & 23.31 & 0.842 & 27.79 & 0.849 & 29.05 & 0.888 & 26.72 & 0.860 \cr
                    \hline
                    \multirow{12}{*}{\centering All-in-One}
                    & All-in-One \cite{LiTC20} & CVPR'20 & 24.71 & 0.898 & 28.33 & 0.882 & 31.12 & 0.927 & 28.05 & 0.902 \cr
                    & TransWeather \cite{ValanarasuYP22} & CVPR'22 & 28.83 & 0.900 & 29.31 & 0.888 & 30.17 & 0.916 & 29.44 & 0.901 \cr
                    & Weather-Diffusion \cite{OzdenizciL23} & TPAMI'23  & 29.72 & 0.922 & 29.58 & 0.894 & 29.66 & 0.923 & 29.65 & 0.913 \cr
                    & PromptIR \cite{PotlapalliZ0K23} & NIPS'23 & 28.75 & 0.897 & 28.99 & 0.871 & 29.98 & 0.915 & 29.24 & 0.894 \cr
                    & APM \cite{abs-2411-16739} & arXiv'24 & 29.22 & 0.910 & 29.56 & 0.890 & 30.76 & 0.910 & 29.85 & 0.903 \cr
                    & DiffUIR-L \cite{ZhengWYZH024} & CVPR'24 & 30.89 & 0.923 & 30.64 & 0.908 & \textbf{31.90}  & 0.937 & 31.14 & 0.923 \cr
                    & MoFME \cite{zhang2024efficient} & AAAI'24 & 28.10 & 0.938 & 28.84 & 0.893 & 29.03 & 0.935 & 28.66 & 0.922 \cr
                    & MWFormer \cite{MWFormer} & TIP'24 & 29.07 & 0.901 & 30.05 & 0.899 & 31.09 & 0.922 & 30.07 & 0.907 \cr
                    & GridFormer \cite{wang2024gridformer} & IJCV'24 & 30.48 & 0.931 & 30.78 & \textbf{0.917} & 31.02 & 0.930 & 30.76 & 0.926 \cr
                    & AdaIR \cite{cui2025adair} & ICLR'25 & 31.01 & 0.923 & 30.57 & 0.902 & 30.31 & 0.917 & 30.63 & 0.914 \cr
                    & M2Restore & Ours & \textbf{32.01} & \textbf{0.955} & \textbf{31.20} & \underline{0.911} & \underline{31.73} & \textbf{0.943} & \textbf{31.65} & \textbf{0.936} \cr
				\bottomrule
			\end{tabular}
		}
	\end{threeparttable}
\end{table*}

\section{Experiments}
In this section, we comprehensively evaluate our M2Restore across three representative image restoration tasks: rain removal, snow removal, and raindrop removal. We begin by detailing the implementation setup, benchmark datasets, and evaluation protocols employed in our experiments. Subsequently, we present quantitative comparisons against state-of-the-art methods, followed by in-depth analytical studies to validate the effectiveness and rationality of our approach.

\subsection{Experiment Setting}
\subsubsection{Implementation Details} We implement our method using the Pytorch framework, and all experiments are conducted on an NVIDIA RTX 3090 GPU. The network is optimized using the Adam optimizer with an initial learning rate of 0.0002. The training process spans 200 complete epochs with a mini-batch size of 2. To mitigate GPU memory constraints, we adopt a gradient accumulation strategy with a step size of 4, effectively simulating a batch size of 8. During training, input images are cropped into patches of size 224$\times$224 pixels. Data augmentation is applied by randomly flipping these patches both horizontally and vertically.

\subsubsection{Datasets} Our M2Restore is trained on the All-weather \cite{LiTC20} dataset, a comprehensive collection of images affected by various adverse weather conditions. The training follows the configuration established by the All-in-One Network benchmark \cite{LiTC20}, ensuring fair and consistent performance comparisons. The All-weather dataset comprises three primary components: (1) 9,000 synthetically generated snow images from the Snow100K dataset \cite{LiuJHH18}, (2) 1,069 real-world raindrop-occluded images from the Raindrop dataset \cite{QianTYS018}, and (3) 9,000 artificially synthesized fog-rain composite images from the Outdoor-Rain dataset \cite{LiCT19}. Specifically, The Snow100K dataset \cite{LiuJHH18} includes images with computationally rendered snow particle accumulations, providing diverse snow degradation patterns. The Raindrop dataset \cite{QianTYS018} contributes genuine photographic samples featuring raindrop artifacts on camera lenses. The Outdoor-Rain dataset \cite{LiCT19} enhances the training diversity by combining physically simulated rainfall streaks with atmospheric haze dispersion effects, yielding complex weather-induced degradations. This comprehensive collection, referred to as the All-weather dataset, integrates both synthetic and real-world degradation patterns caused by snow, rain, and haze, facilitating robust feature learning for all-in-one image restoration. Additionally, the generalization capability of M2Restore is evaluated on the RainDS-real \cite{Quan0L021} dataset, which comprises real-world rainy scenes.

\subsubsection{Comparison Methods and Evaluation Metrics } We compare our approach against state-of-the-art (SOTA) methods, including both general and all-in-one image restoration models. For general methods, we evaluate six representative models: MPRNet \cite{ZamirA0HK0021}, SwinIR \cite{LiangCSZGT21}, NAFNet \cite{ChenCZS22}, Uformer \cite{WangCBZLL22}, Restormer \cite{ZamirA0HK022}, and GRL \cite{LiFXDRTG23}. For all-in-one restoration, we include comparisons with pioneering and recent approaches: All-in-One  \cite{LiTC20}, TransWeather \cite{ValanarasuYP22}, Weather-Diffusion \cite{OzdenizciL23}, PromptIR \cite{PotlapalliZ0K23}, APM \cite{abs-2411-16739},DiffUIR-L \cite{ZhengWYZH024}, MoFME \cite{zhang2024efficient}, MWFormer \cite{MWFormer}, GridFormer\cite{wang2024gridformer} and the recently proposed AdaIR \cite{cui2025adair}. Restoration performance is assessed using Peak Signal-to-Noise Ratio (PSNR) and Structural Similarity Index (SSIM) metrics, calculated on the RGB channels. Higher PSNR and SSIM values indicate better perceptual and structural fidelity of the reconstructed images.

\begin{figure*}[htbp] 
    \centering
    \includegraphics[width=0.97\linewidth]{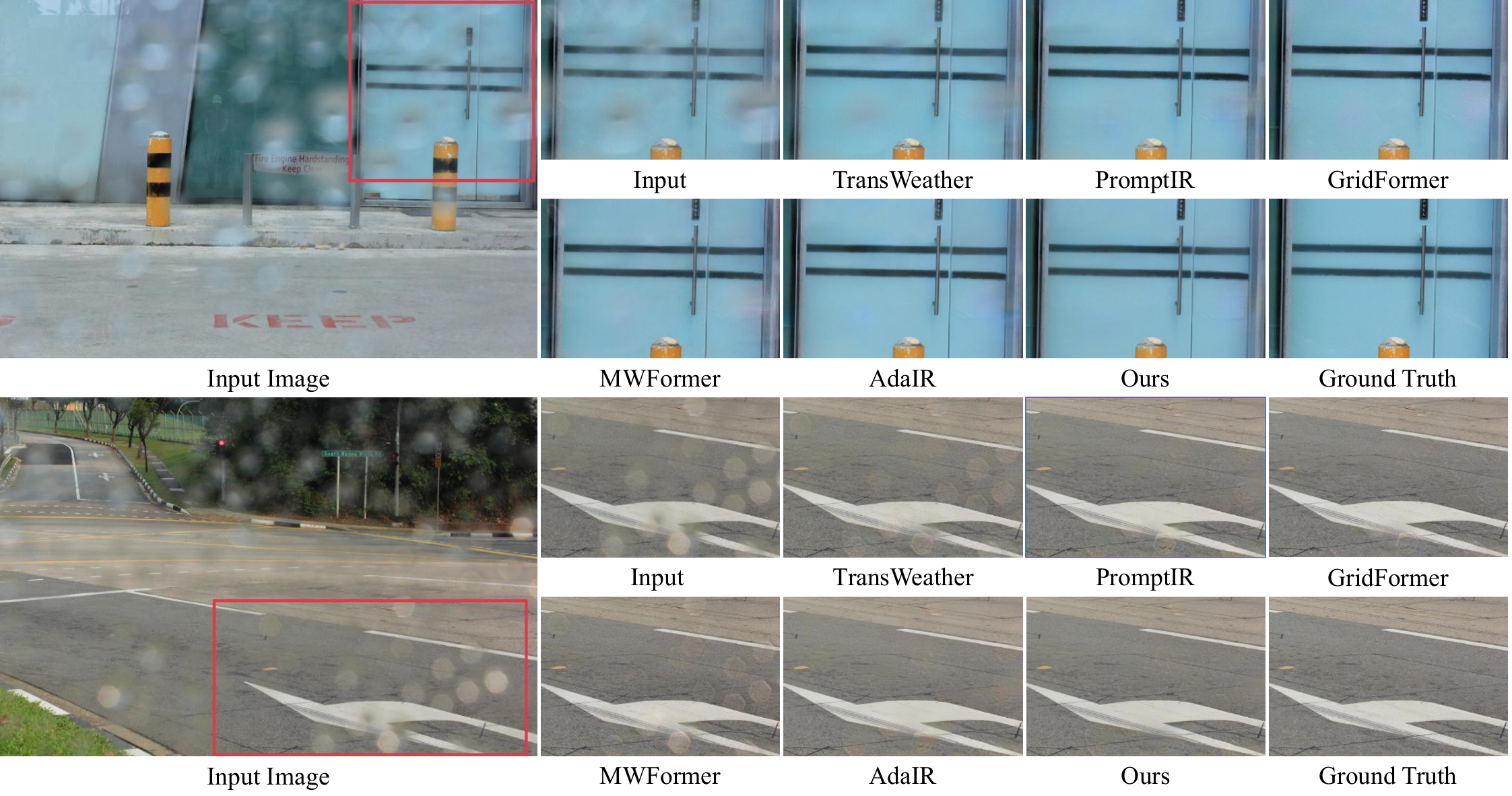}
	\caption{
Qualitative Comparison of different methods on the Raindrop \cite{QianTYS018} dataset. In contrast to other approaches, our M2Restore achieves superior restoration by thoroughly removing raindrop-induced blur and preserving fine image details.
 	}
 	\label{fig:fig4}
 \end{figure*}
 

\begin{figure*}[!h] 
    \centering
    \includegraphics[width=0.97\linewidth]{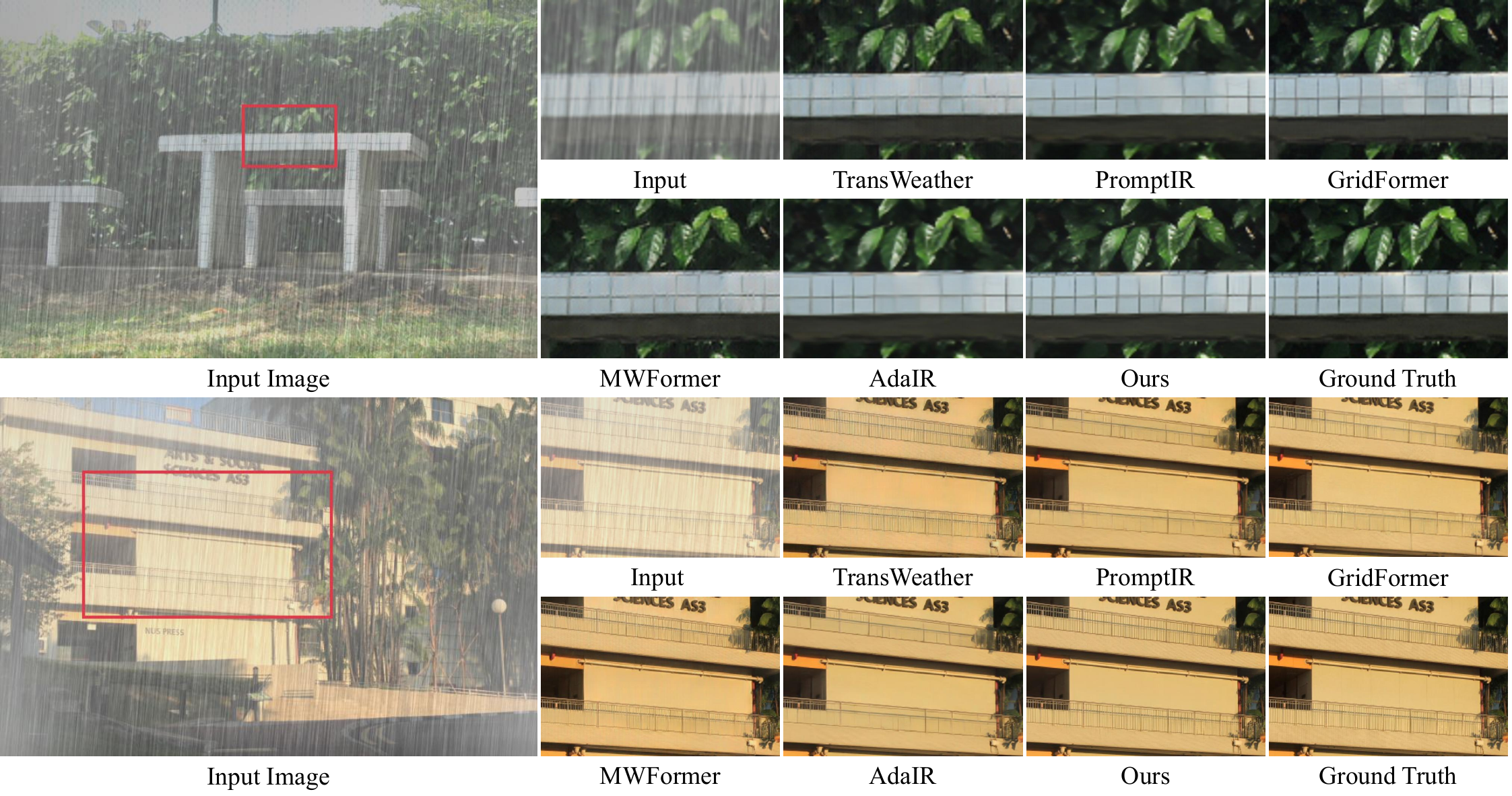}
	\caption{
Qualitative comparison of different methods on the Outdoor-Rain \cite{LiCT19} dataset (Rain + Fog). Compared to other approaches, our M2Restore demonstrates superior capability in preserving intricate structural details while ensuring high perceptual fidelity.}
 	\label{fig:fig5}
 \end{figure*}

\begin{figure*}[t] 
    \centering
    \includegraphics[width=0.97\linewidth]{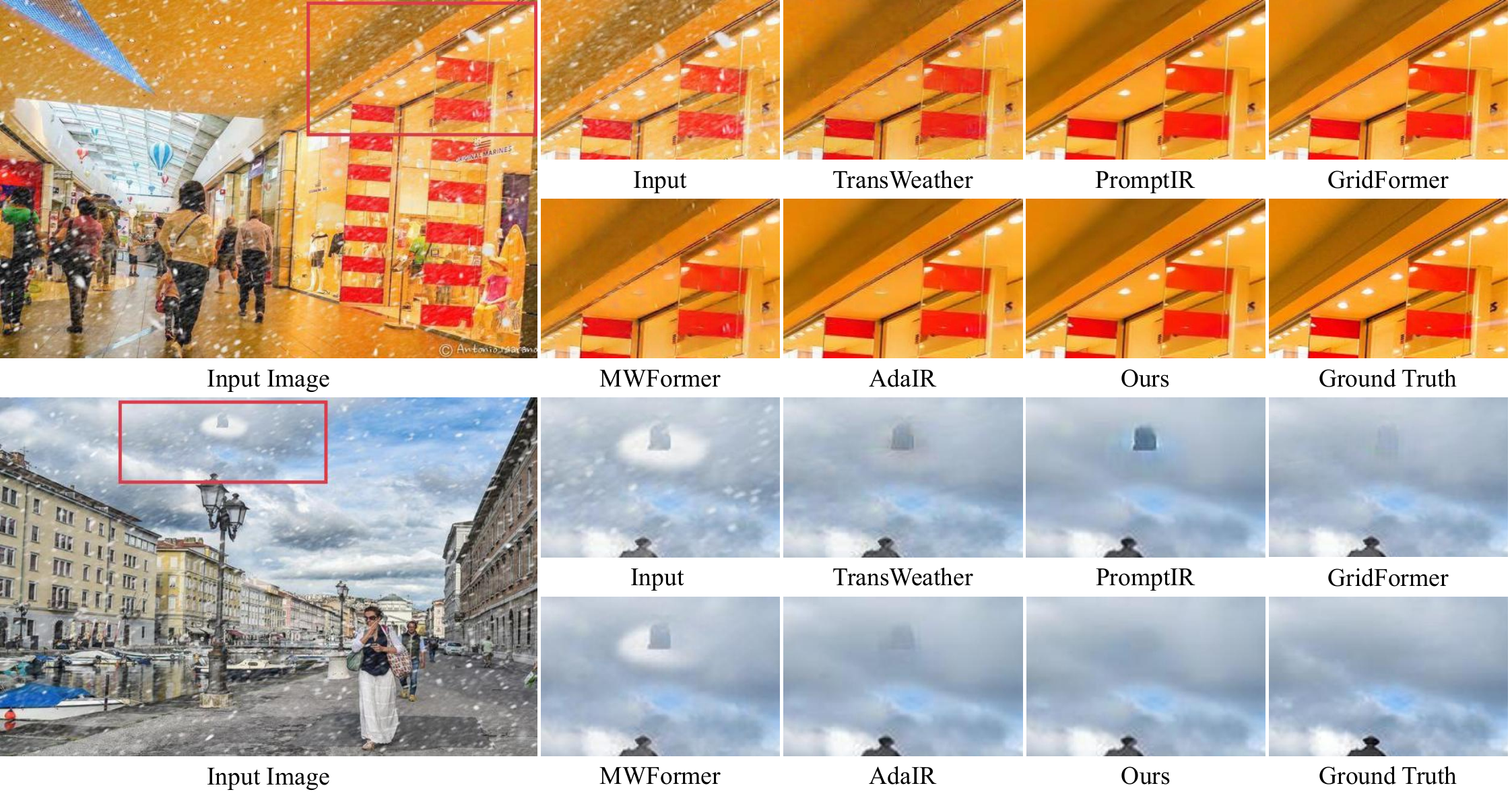}
	\caption{
Qualitative comparison of different methods on the Snow100K-L  \cite{LiuJHH18} dataset. Compared to other approaches, our M2Restore more effectively differentiates light sources from snow and excels at removing large snow particles.
 	}
 	\label{fig:fig6}
 \end{figure*}

\begin{figure*}[!h] 
    \centering
    \includegraphics[width=0.97\linewidth]{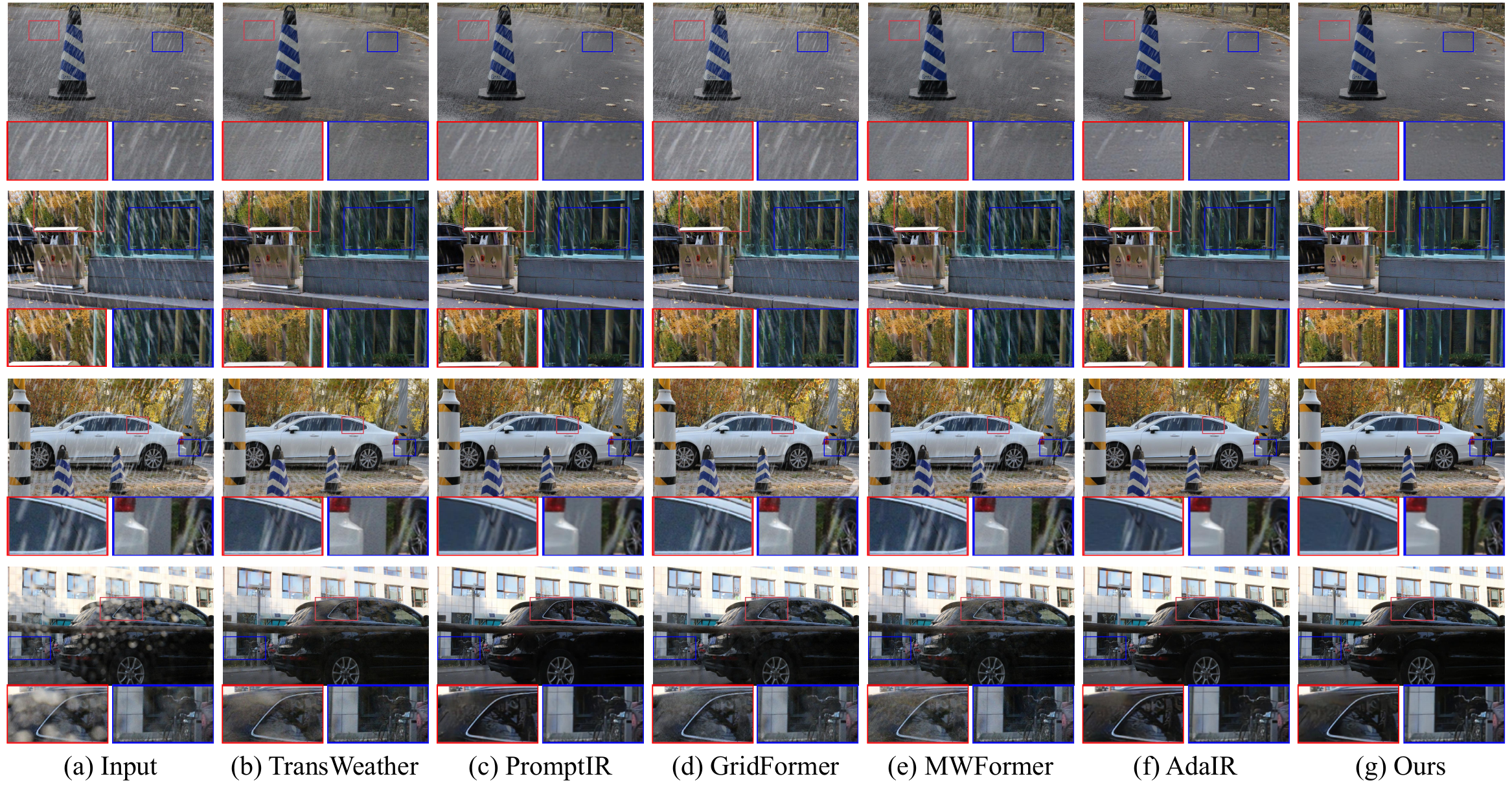}
	\caption{
    Qualitative comparison of different methods on real-world images form the RainDS-real \cite{Quan0L021} dataset. As illustrated, our M2Restore produces reconstructed images with superior visual fidelity and markedly reduced rain artifacts. This improvement stems from its expert scheduling strategy combined with a global-local dual restoration architecture.
 	}
 	\label{fig:fig7}
 \end{figure*}
 
\subsection{Comparison with State-of-the-art Methods}
\subsubsection{Quantitative Comparison}
We benchmark our M2Restore against eleven image restoration methods using two quantitative metrics(PSNR and SSIM). As summarized in Table \ref{Tab:tab1}, our method achieves the best average performance across all three benchmark datasets, demonstrating robust generalization across diverse degradation types. Specifically, although DiffUIR-L attains the highest PSNR of 31.90 dB on the Raindrop dataset, M2Restore achieves a closely comparably score of 31.73 dB, consistently ranking second and demonstrating competitive performance. On the Snow100K-L dataset, GridFormer obtains the highest SSIM of 0.917; however, its PSNR of 30.78 dB falls notably short compared to M2Restore’s 31.20 dB, indicating a trade-off between structural fidelity and reconstruction accuracy. Moreover, GridFormer exhibits significantly weaker performance on the Outdoor-Rain dataset, underscoring its limited generalization to complex or mixed degradations. Overall, these results affirm that M2Restore delivers the most balanced performance across all datasets, effectively managing diverse and challenging degradation types with superior consistency and robustness.

\subsubsection{Qualitative Comparison}
To further evaluate the effectiveness of M2Restore, we conduct comprehensive visual assessments across all benchmark datasets. As illustrated in Fig. \ref{fig:fig4}, which displays two samples from the raindrop test set, M2Restore demonstrates superior performance in handling large raindrops and heavily occluded regions compared to other methods. Notably, in the first row (glass door surfaces) and at the indication mark edges in subsequent rows, competing approaches such as PromptIR, MWFormer, GridFormer, and AdaIR introduce visible artifacts, including shadow effects and hallucinated details. In contrast, M2Restore delivers visually compelling results, accurately restoring complex textures and fine structures while effectively mitigating blurring and shadow artifacts.

Further evaluation on the Outdoor-Rain \cite{LiCT19} (Rain + Fog) dataset, depicted in Fig. \ref{fig:fig5}, reveals M2Restore’s strong capabilities in both luminance recovery and detail preservation. Specifically, reconstructions produced by other methods tend to suffer from excessive smoothing during rain-fog removal, leading to a loss of structural detail (e.g., the gaps between tiles in the first sample and the metal railings in the second). In comparison, our M2Restore successfully restores such details, maintaining the integrity of the scene.

Fig. \ref{fig:fig6} illustrates snow removal results on the Snow100k-L \cite{LiuJHH18} test set. While all six methods deliver generally satisfactory results, M2Restore consistently outperforms others in more challenging scenarios: 1) Circular light sources (rows 1–2) are mistakenly removed as snow artifacts by other methods, whereas M2Restore preserves them accurately; 2) For larger snow particles (rows 3–4), residual halos remain in the results from MWFormer, PromptIR, TransWeather, GridFormer, and AdaIR, while M2Restore achieves clean removal without introducing distortions. These visual comparisons clearly demonstrate M2Restore’s advanced restoration capabilities across diverse and complex weather degradation scenarios.

\subsubsection{Comparison on Real-World Images} 
To assess the practical effectiveness of our method under real-world conditions,  we conduct comparative experiments on the RainDS-real \cite{Quan0L021} dataset, which contains complex rain streak and raindrop patterns captured in real-world outdoor scenes. As illustrated in Fig. \ref{fig:fig7}, we analyze restoration results across four representative groups involving challenging scenarios: road surfaces (first row), building glass (second row), vehicle windows (third row), and wall surfaces (fourth row). M2Restore consistently outperforms existing methods, achieving visually cleaner outputs with more accurate structural preservation and significantly fewer artifacts. The advantages of our approach are especially evident in the following two observations: 1) In the first-row road surfaces (highlighted by two zoomed-in regions), second-row building glass (blue box), and third-row column area (blue box), comparative methods struggle with residual rain-induced blur and incomplete streak removal. M2Restore, by contrast, leverages multi-scale feature fusion to produce clearer, more complete restorations; 2) In the fourth-row wall surface, other methods show residual blur caused by raindrop occlusion (e.g., red box on the window) and introduce visible artifacts (e.g., lower-left region in the blue box). M2Restore delivers artifact-free, high-fidelity reconstructions with enhanced sharpness and well-preserved textures. These findings underscore the robustness of our method in handling real-world degradations, affirming its superior generalization capability and practical effectiveness in diverse image restoration scenarios.


\subsection{Ablation Study}
The proposed M2Restore demonstrates strong performance across all-in-one image restoration tasks. To rigorously evaluate the contribution of each individual component, we conduct comprehensive ablation studies on three representative restoration scenarios: rain/fog removal, raindrop removal, and snow removal. Quantitative evaluations, averaged over these three tasks, are presented in Table \ref{Tab:2}. M2Restore integrates three key modules: DDER, MCDB, and DGF, each playing a pivotal role in the overall restoration pipeline. All experimental variants are trained under identical hyperparameter configurations to ensure fair comparison and reliable conclusions.

\textbf{Effectiveness of DDER.} The Dynamic Degradation-aware Expert Router (DDER) facilitates task-specific expert activation via cross-attention between degradation types and image content. To evaluate its contribution, we remove all DDER modules from the encoder and substitute them with vanilla Transformer layers. As exhibited in Table \ref{Tab:2}, this leads to a notable performance drop, highlighting DDER’s critical role in encoding degradation-specific contextual correlations. The performance drop can be attributed to the absence of (1) adaptively route information based on degradation type, and (2) modulate feature representations dynamically through expert selection.

\textbf{Effectiveness of DGF.} The Dynamic Gated Fusion (DGF) module acts as a spatial-channel attention mechanism, emphasizing high-frequency details and degradation-sensitive regions. Disabling DGF by replacing its attention-based fusion with static aggregation significantly weakens performance across all benchmarks. This validates DGF’s role in preserving fine-grained structures and ensuring adaptive feature prioritization, particularly in scenarios involving complex or overlapping degradations.

\textbf{Effectiveness of MCDB.} The Multi-contextual Collaborative Decoder Block (MCDB) is designed to re-integrate high-frequency edges and degradation-sensitive features, based on the encoder’s outputs.  To evaluate its role, we retain only the DDER component while replacing both MCDB and DGF in the bottleneck layer with a standard Transformer module. As reported in Table~\ref{Tab:2}, this configuration results in noticeable performance degradation, highlighting MCDB’s complementary role in strengthening the network's representation capability for fine-grained restoration tasks.

\begin{table}[t]
	\centering
	\footnotesize
	\caption{Ablation analysis of different module combinations on the All-weather dataset. Our complete model attains the best performance}
	\label{Tab:2}
	\begin{threeparttable}
		\centering
		\setlength{\tabcolsep}{3mm}{
			\begin{tabular}{ccc|cc}
				\toprule
				\multirow{1}{*}{DDER}&
				\multirow{1}{*}{DGF}&
				\multirow{1}{*}{MCDB}&
				\multirow{1}{*}{PSNR}&
				\multirow{1}{*}{SSIM}\cr
				\midrule
                \checkmark & w/o & \checkmark  & 31.21 & 0.9309 \cr
                w/o &\checkmark & \checkmark & 31.13 & 0.9304 \cr
				\checkmark & w/o & w/o  & 31.35& 0.9305 \cr
                \checkmark &\checkmark & \checkmark & 31.65 & 0.9362 \cr
				\bottomrule
			\end{tabular}
		}
	\end{threeparttable}
\end{table}

\begin{table}[t]
	\centering
	\footnotesize
	\caption{Average runtime comparison among state-of-the-art restoration methods. Red and blue indicate the best and second-best results, respectively}
	\label{Tab:3}
	\begin{threeparttable}
		\centering
		\setlength{\tabcolsep}{3mm}{
			\begin{tabular}{ccc}
				\toprule
				\multirow{1}{*}{Method}&
				\multirow{1}{*}{Publication}&
				\multirow{1}{*}{Inference Time ($s$)}\cr
				\midrule
                All-in-one & CVPR'20 & 0.33 \cr
                MPRNet & CVPR'21 & 0.18 \cr
                SwinIR & ICCV'21 & 0.27 \cr
                Restormer & CVPR'22 & 0.18 \cr
                WeatherDiff & TPAMI'23 & 58.16 \cr
                PromptIR & NIPS'23 & 0.44 \cr
                APM & arXiv'24 & 0.33 \cr
                MoFME & AAAI'24 & 0.21 \cr
                GridFormer & IJCV'24 & 1.43 \cr
                MWFormer  & TIP'24 & 0.33 \cr
                AdaIR & ICLR'25 & \color{red}0.13 \cr
                M2Restore & Ours & \color{blue}0.17 \cr
				\bottomrule
			\end{tabular}
		}
	\end{threeparttable}
\end{table}

\subsection{Efficiency Analysis}
Given the importance of computational efficiency in real-world computer vision applications, we evaluate the inference speed of several image restoration algorithms. Table \ref{Tab:3} reports the average runtime (in seconds) for M2Restore and 11 representative state-of-the-art methods, all evaluated using a consistent input resolution of 224$\times$224 pixels on an NVIDIA RTX 3090 GPU. It can be observed that M2Restore achieves a competitive inference speed, requiring only 0.17 seconds on average to restore a degraded image. This makes it the second-fastest model among the compared methods. The high efficiency of M2Restore can be attributed to its hybrid expert architecture, which leverages sparse matrix operations to reduce the computational burden while preserving performance. This design enables effective degradation handling with significantly fewer parameters compared to dense models.

\begin{figure}[t] 
    \centering
    \includegraphics[width=1.0\linewidth]{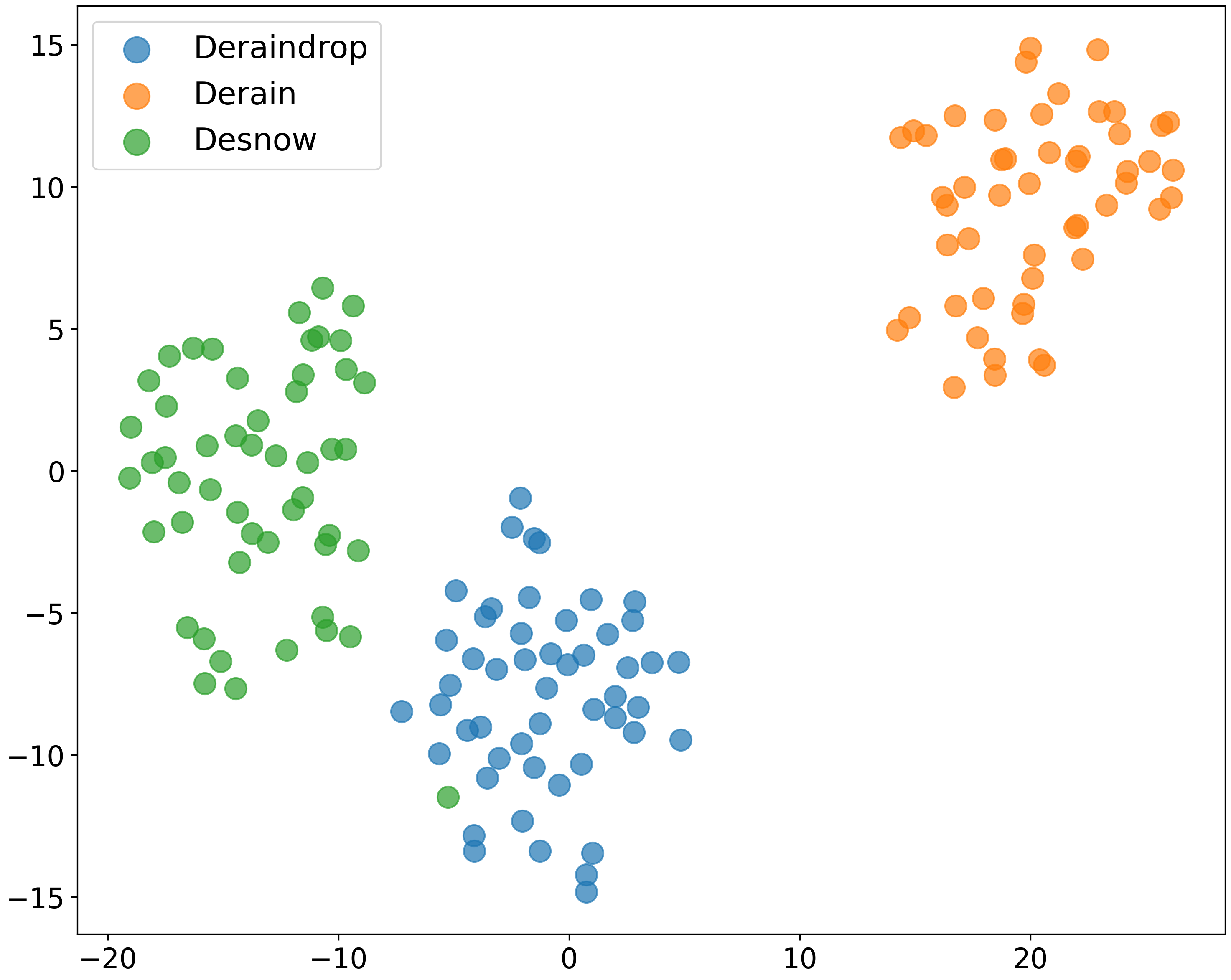}
	\caption{
    Visualization of degradation-aware expert router (DDER) weights in M2Restore via t-SNE
    }
 	\label{fig:fig8}
 \end{figure}

\subsection{Discussions}
To further examine the expert selection dynamics within M2Restore, we conduct an in-depth analysis using 150 randomly sampled images from the All-weather test set, equally distributed across three degradation types: raindrop occlusion (50 samples), rain streaks (50 samples), and snow accumulation (50 samples). We apply t-SNE visualization with PCA initialization (perplexity = 30) to project the expert weighting vectors produced by the DDER module into a 2D latent space. The results reveal clearly separated clusters corresponding to different degradation types. Quantitative metrics further support this observation: the average Silhouette score reaches 0.25, indicating moderate inter-class separability, while the intra-cluster cohesion (mean Euclidean distance: 1.27 ± 0.43) is notably lower than the inter-cluster distance (1.78 ± 0.34). As illustrated in Fig. \ref{fig:fig8}, weather-specific samples form compact and coherent clusters, while cross-degradation boundaries remain distinct. This emergent geometric structure demonstrates that DDER effectively learns to encode degradation-aware representations and dynamically selects relevant experts based on input conditions.


\section{Conclusion}
In this work, we introduce M2Restore, an all-in-one image restoration framework that seamlessly integrates a Mixture-of-Experts paradigm with a Mamba-CNN fusion network. To achieve high-quality restoration of degraded images, we design a Dynamic Degradation-Aware Expert Router (DDER), which enables M2Restore to adaptively select the most pertinent experts by jointly attending to both image content and degradation patterns. Leveraging Mamba's remarkable ability to capture long-range dependencies with linear computational complexity, we construct a Mamba-CNN Dual-Branch hybrid architecture (MCDB) that effectively combines the complementary strengths of Mamba and CNNs for joint global-local feature representation and fine-grained reconstruction. Furthermore, a Dynamic Gated Feature Fusion (DGF) strategy is incorporated into MCDB to adaptively modulate the contributions of the Mamba and CNN branches, thereby enhancing the network’s sensitivity to degradation-sensitive regions and promoting precise structural and textural restoration. Extensive experiments on both synthetic and real-world datasets demonstrate the superiority of the proposed method.

\bibliographystyle{IEEEtran}
\bibliography{reference}

\begin{thebibliography}{10}
\providecommand{\url}[1]{#1}
\csname url@samestyle\endcsname
\providecommand{\newblock}{\relax}
\providecommand{\bibinfo}[2]{#2}
\providecommand{\BIBentrySTDinterwordspacing}{\spaceskip=0pt\relax}
\providecommand{\BIBentryALTinterwordstretchfactor}{4}
\providecommand{\BIBentryALTinterwordspacing}{\spaceskip=\fontdimen2\font plus
\BIBentryALTinterwordstretchfactor\fontdimen3\font minus \fontdimen4\font\relax}
\providecommand{\BIBforeignlanguage}[2]{{%
\expandafter\ifx\csname l@#1\endcsname\relax
\typeout{** WARNING: IEEEtran.bst: No hyphenation pattern has been}%
\typeout{** loaded for the language `#1'. Using the pattern for}%
\typeout{** the default language instead.}%
\else
\language=\csname l@#1\endcsname
\fi
#2}}
\providecommand{\BIBdecl}{\relax}
\BIBdecl

\bibitem{MengWDXP13}
G.~Meng, Y.~Wang, J.~Duan, S.~Xiang, and C.~Pan, ``Efficient image dehazing with boundary constraint and contextual regularization,'' in \emph{Proc. IEEE Int. Conf. Comput. Vis. (ICCV)}, 2013, pp. 617--624.

\bibitem{TimofteDG13}
R.~Timofte, V.~D. Smet, and L.~V. Gool, ``Anchored neighborhood regression for fast example-based super-resolution,'' in \emph{Proc. IEEE Int. Conf. Comput. Vis. (ICCV)}, 2013, pp. 1920--1927.

\bibitem{ZhuMS15}
Q.~Zhu, J.~Mai, and L.~Shao, ``A fast single image haze removal algorithm using color attenuation prior,'' \emph{{IEEE} Trans. Image Process.}, vol.~24, no.~11, pp. 3522--3533, 2015.

\bibitem{KimK10}
K.~I. Kim and Y.~Kwon, ``Single-image super-resolution using sparse regression and natural image prior,'' \emph{{IEEE} Trans. Pattern Anal. Mach. Intell.}, vol.~32, no.~6, pp. 1127--1133, 2010.

\bibitem{He0T11}
K.~He, J.~Sun, and X.~Tang, ``Single image haze removal using dark channel prior,'' \emph{{IEEE} Trans. Pattern Anal. Mach. Intell.}, vol.~33, no.~12, pp. 2341--2353, 2011.

\bibitem{ChenMSINR2024}
X.~Chen, J.~Pan, and J.~Dong, ``Bidirectional multi-scale implicit neural representations for image deraining,'' in \emph{Proc. IEEE/CVF Conf. Comput. Vis. Pattern Recognit. (CVPR)}, 2024, pp. 25\,627--25\,636.

\bibitem{wang2024ucl}
Y.~Wang, X.~Yan, F.~L. Wang, H.~Xie, W.~Yang, X.-P. Zhang, J.~Qin, and M.~Wei, ``Ucl-dehaze: toward real-world image dehazing via unsupervised contrastive learning,'' \emph{{IEEE} Trans. Image Process.}, vol.~33, pp. 1361--1374, 2024.

\bibitem{Wang2023SmartAssign}
Y.~Wang, C.~Ma, and J.~Liu, ``Smartassign: Learning {A} smart knowledge assignment strategy for deraining and desnowing,'' in \emph{Proc. IEEE/CVF Conf. Comput. Vis. Pattern Recognit. (CVPR)}, 2023, pp. 3677--3686.

\bibitem{Wang2025APANet}
C.~Wang, T.~Yan, W.~Huang, X.~Chen, K.~Xu, and X.~Chang, ``Apanet: Asymmetrical parallax attention network for efficient stereo image deraining,'' \emph{{IEEE} Trans. Computational Imaging}, vol.~11, pp. 101--115, 2025.

\bibitem{Quan2023InvDSNet}
Y.~Quan, X.~Tan, Y.~Huang, Y.~Xu, and H.~Ji, ``Image desnowing via deep invertible separation,'' \emph{{IEEE} Trans. Circuits Syst. Video Technol.}, vol.~33, no.~7, pp. 3133--3144, 2023.

\bibitem{LiTC20}
R.~Li, R.~T. Tan, and L.~Cheong, ``All in one bad weather removal using architectural search,'' in \emph{Proc. IEEE/CVF Conf. Comput. Vis. Pattern Recognit. (CVPR)}, 2020, pp. 3172--3182.

\bibitem{Ai2024MPer-ceiver}
Y.~Ai, H.~Huang, X.~Zhou, J.~Wang, and R.~He, ``Multimodal prompt perceiver: Empower adaptiveness, generalizability and fidelity for all-in-one image restoration,'' in \emph{Proc. IEEE/CVF Conf. Comput. Vis. Pattern Recognit. (CVPR)}, 2024, pp. 25\,432--25\,444.

\bibitem{Zhu2024MWDAT}
R.~Zhu, M.~Wu, X.~Xiong, X.~Zhu, and Y.~Fan, ``Multi-weather degradation-aware transformer for image restoration,'' in \emph{Proc. {IEEE} Int. Conf. Acoustics}, 2024, pp. 3765--3769.

\bibitem{ParkLC23}
D.~Park, B.~H. Lee, and S.~Y. Chun, ``All-in-one image restoration for unknown degradations using adaptive discriminative filters for specific degradations,'' in \emph{Proc. IEEE/CVF Conf. Comput. Vis. Pattern Recognit. (CVPR)}, 2023, pp. 5815--5824.

\bibitem{ZhangDMW25}
S.~Zhang, Q.~Dong, W.~Mao, and Z.~Wang, ``A unified accelerator for all-in-one image restoration based on prompt degradation learning,'' \emph{{IEEE} Trans. Circuits Syst. {I} Regul. Pap.}, vol.~72, no.~3, pp. 1282--1295, 2025.

\bibitem{10453462}
Z.~Tan, Y.~Wu, Q.~Liu, Q.~Chu, L.~Lu, J.~Ye, and N.~Yu, ``Exploring the application of large-scale pre-trained models on adverse weather removal,'' \emph{IEEE Trans. Image Process.}, vol.~33, pp. 1683--1698, 2024.

\bibitem{JiangZXG24}
Y.~Jiang, Z.~Zhang, T.~Xue, and J.~Gu, ``Autodir: Automatic all-in-one image restoration with latent diffusion,'' in \emph{Proc. Eur. Conf. Comput. Vis. (ECCV)}, vol. 15098, 2024, pp. 340--359.

\bibitem{MWFormer}
R.~Zhu, Z.~Tu, J.~Liu, A.~C. Bovik, and Y.~Fan, ``Mwformer: Multi-weather image restoration using degradation-aware transformers,'' \emph{IEEE Trans. Image Process.}, vol.~33, pp. 6790--6805, 2024.

\bibitem{LiangCSZGT21}
J.~Liang, J.~Cao, G.~Sun, K.~Zhang, L.~V. Gool, and R.~Timofte, ``Swinir: Image restoration using swin transformer,'' in \emph{Proc. IEEE Int. Conf. Comput. Vis. (ICCV) Workshops}, 2021, pp. 1833--1844.

\bibitem{ZamirA0HK022}
S.~W. Zamir, A.~Arora, S.~Khan, M.~Hayat, F.~S. Khan, and M.~Yang, ``Restormer: Efficient transformer for high-resolution image restoration,'' in \emph{Proc. IEEE/CVF Conf. Comput. Vis. Pattern Recognit. (CVPR)}, 2022, pp. 5718--5729.

\bibitem{cui2025adair}
Y.~Cui, S.~W. Zamir, S.~Khan, A.~Knoll, M.~Shah, and F.~S. Khan, ``Ada{IR}: Adaptive all-in-one image restoration via frequency mining and modulation,'' in \emph{Proc. Int. Conf. Learn. Represent. (ICLR)}, 2025.

\bibitem{Gu2023Mamba}
A.~Gu and T.~Dao, ``Mamba: Linear-time sequence modeling with selective state spaces,'' \emph{arXiv preprint arXiv:2312.00752}, 2023.

\bibitem{jiang2024multi}
A.~Jiang, H.~Chen, Z.~Chen, J.~Ye, and M.~Wang, ``Multi-dimensional visual prompt enhanced image restoration via mamba-transformer aggregation,'' \emph{arXiv preprint arXiv:2412.15845}, 2024.

\bibitem{JacobsJNH91}
R.~A. Jacobs, M.~I. Jordan, S.~J. Nowlan, and G.~E. Hinton, ``Adaptive mixtures of local experts,'' \emph{Neural Comput.}, vol.~3, no.~1, pp. 79--87, 1991.

\bibitem{YuWDTL22}
K.~Yu, X.~Wang, C.~Dong, X.~Tang, and C.~C. Loy, ``Path-restore: Learning network path selection for image restoration,'' \emph{{IEEE} Trans. Pattern Anal. Mach. Intell.}, vol.~44, no.~10, pp. 7078--7092, 2022.

\bibitem{RadfordKHRGASAM21}
A.~Radford, J.~W. Kim, C.~Hallacy, A.~Ramesh, G.~Goh, S.~Agarwal, G.~Sastry, A.~Askell, P.~Mishkin, J.~Clark, G.~Krueger, and I.~Sutskever, ``Learning transferable visual models from natural language supervision,'' in \emph{Proc. Int. Conf. Mach. Learn. (ICML)}, vol. 139, 2021, pp. 8748--8763.

\bibitem{xiong2025da2diffexploringdegradationawareadaptive}
J.~Xiong, X.~Yan, Y.~Wang, W.~Zhao, X.-P. Zhang, and M.~Wei, ``Da2diff: Exploring degradation-aware adaptive diffusion priors for all-in-one weather restoration,'' \emph{arXiv preprint arXiv:2504.05135}, 2025.

\bibitem{ai2024lora}
Y.~Ai, H.~Huang, and R.~He, ``Lora-ir: Taming low-rank experts for efficient all-in-one image restoration,'' \emph{arXiv preprint arXiv:2410.15385}, 2024.

\bibitem{LuoG0SS24}
Z.~Luo, F.~K. Gustafsson, Z.~Zhao, J.~Sj{\"{o}}lund, and T.~B. Sch{\"{o}}n, ``Controlling vision-language models for multi-task image restoration,'' in \emph{Proc. Int. Conf. Learn. Represent. (ICLR)}, 2024.

\bibitem{BuadesCM05}
A.~Buades, B.~Coll, and J.~Morel, ``A non-local algorithm for image denoising,'' in \emph{Proc. {IEEE} Conf. Comput. Vis. Pattern Recognit. (CVPR)}, 2005, pp. 60--65.

\bibitem{olshausen1996emergence}
B.~A. Olshausen and D.~J. Field, ``Emergence of simple-cell receptive field properties by learning a sparse code for natural images,'' \emph{Nature}, vol. 381, no. 6583, pp. 607--609, 1996.

\bibitem{ZhangZCM017}
K.~Zhang, W.~Zuo, Y.~Chen, D.~Meng, and L.~Zhang, ``Beyond a gaussian denoiser: Residual learning of deep {CNN} for image denoising,'' \emph{{IEEE} Trans. Image Process.}, vol.~26, no.~7, pp. 3142--3155, 2017.

\bibitem{TianZWXZCL20}
C.~Tian, R.~Zhuge, Z.~Wu, Y.~Xu, W.~Zuo, C.~Chen, and C.~Lin, ``Lightweight image super-resolution with enhanced {CNN},'' \emph{Knowl. Based Syst.}, vol. 205, p. 106235, 2020.

\bibitem{DosovitskiyB0WZ21}
A.~Dosovitskiy, L.~Beyer, A.~Kolesnikov, D.~Weissenborn, X.~Zhai, T.~Unterthiner, M.~Dehghani, M.~Minderer, G.~Heigold, S.~Gelly, J.~Uszkoreit, and N.~Houlsby, ``An image is worth 16x16 words: Transformers for image recognition at scale,'' in \emph{Proc. Int. Conf. Learn. Represent. (ICLR)}, 2021.

\bibitem{GuoLDORX24}
H.~Guo, J.~Li, T.~Dai, Z.~Ouyang, X.~Ren, and S.~Xia, ``Mambair: {A} simple baseline for image restoration with state-space model,'' in \emph{Proc. Eur. Conf. Comput. Vis. (ECCV)}, vol. 15076, 2024, pp. 222--241.

\bibitem{LiLHW0022}
B.~Li, X.~Liu, P.~Hu, Z.~Wu, J.~Lv, and X.~Peng, ``All-in-one image restoration for unknown corruption,'' in \emph{Proc. IEEE/CVF Conf. Comput. Vis. Pattern Recognit. (CVPR)}, 2022, pp. 17\,431--17\,441.

\bibitem{PotlapalliZ0K23}
V.~Potlapalli, S.~W. Zamir, S.~H. Khan, and F.~S. Khan, ``Promptir: Prompting for all-in-one image restoration,'' in \emph{Proc. Annu. Conf. Neural Inf. Process. Syst. (NIPS))}, 2023.

\bibitem{ArarVHAFPCS24}
M.~Arar, A.~Voynov, A.~Hertz, O.~Avrahami, S.~Fruchter, Y.~Pritch, D.~Cohen{-}Or, and A.~Shamir, ``{PALP:} prompt aligned personalization of text-to-image models,'' in \emph{Proc. {SIGGRAPH} Asia}, 2024, pp. 5:1--5:11.

\bibitem{guo2022dehamer}
C.-L. Guo, Q.~Yan, S.~Anwar, R.~Cong, W.~Ren, and C.~Li, ``Image dehazing transformer with transmission-aware 3d position embedding,'' in \emph{Proc. IEEE/CVF Conf. Comput. Vis. Pattern Recognit. (CVPR)}, 2022, pp. 5812--5820.

\bibitem{ChenDLCYL20}
Y.~Chen, X.~Dai, M.~Liu, D.~Chen, L.~Yuan, and Z.~Liu, ``Dynamic convolution: Attention over convolution kernels,'' in \emph{Proc. IEEE/CVF Conf. Comput. Vis. Pattern Recognit. (CVPR)}, 2020, pp. 11\,027--11\,036.

\bibitem{RenLLWGZZC24}
Y.~Ren, X.~Li, B.~Li, X.~Wang, M.~Guo, S.~Zhao, L.~Zhang, and Z.~Chen, ``Moe-diffir: Task-customized diffusion priors for universal compressed image restoration,'' in \emph{Proc. Eur. Conf. Comput. Vis. (ECCV)}, A.~Leonardis, E.~Ricci, S.~Roth, O.~Russakovsky, T.~Sattler, and G.~Varol, Eds., vol. 15067, 2024, pp. 116--134.

\bibitem{abs-2503-15868}
D.~Mandal, S.~Chattopadhyay, G.~Tong, and P.~Chakravarthula, ``Unicorn: Latent diffusion-based unified controllable image restoration network across multiple degradations,'' \emph{arXiv preprint arXiv:2503.15868}, 2025.

\bibitem{abs-2412-20157}
J.~Lin, Z.~Zhang, W.~Li, R.~Pei, H.~Xu, H.~Zhang, and W.~Zuo, ``Unirestorer: Universal image restoration via adaptively estimating image degradation at proper granularity,'' \emph{arXiv preprint arXiv:2412.20157}, 2024.

\bibitem{zhang2025}
X.~Zhang, H.~Zhang, G.~Wang, Q.~Zhang, L.~Zhang, and B.~Du, ``Uniuir: Considering underwater image restoration as an all-in-one learner,'' \emph{arXiv preprint arXiv:2501.12981}, 2025.

\bibitem{zamfir2025}
E.~Zamfir, Z.~Wu, N.~Mehta, Y.~Tan, D.~P. Paudel, Y.~Zhang, and R.~Timofte, ``Complexity experts are task-discriminative learners for any image restoration,'' \emph{arXiv preprint arXiv:2411.18466}, 2025.

\bibitem{kim2020restorings}
S.~Kim, N.~Ahn, and K.-A. Sohn, ``Restoring spatially-heterogeneous distortions using mixture of experts network,'' in \emph{Proc. Asian Conf. Comput. Vis. (ACCV)}, 2020.

\bibitem{abs-2106-09667}
N.~Carlini and A.~Terzis, ``Poisoning and backdooring contrastive learning,'' \emph{arXiv preprint arXiv:2106.09667}, 2021.

\bibitem{abs-2411-18466}
E.~Zamfir, Z.~Wu, N.~Mehta, Y.~Tan, D.~P. Paudel, Y.~Zhang, and R.~Timofte, ``Complexity experts are task-discriminative learners for any image restoration,'' \emph{arXiv preprint arXiv:2411.18466}, 2024.

\bibitem{TangWZ25}
A.~Tang, Y.~Wu, and Y.~Zhang, ``Ramir: Reasoning and action prompting with mamba for all-in-one image restoration,'' \emph{Appl. Intell.}, vol.~55, no.~4, p. 258, 2025.

\bibitem{MatIR}
J.~Wen, W.~Hou, L.~V. Gool, and R.~Timofte, ``Matir: {A} hybrid mamba-transformer image restoration model,'' \emph{arXiv preprint arXiv:2501.18401}, 2025.

\bibitem{abs-2404-11778}
R.~Deng and T.~Gu, ``Cu-mamba: Selective state space models with channel learning for image restoration,'' \emph{arXiv preprint arXiv:2404.11778}, 2024.

\bibitem{PengGS25}
Y.~Peng, G.~Gao, and C.~Shi, ``Learning a multi-scale vision mamba for weather-degraded remote sensing image restoration,'' \emph{Signal Image Video Process.}, vol.~19, no.~4, p. 274, 2025.

\bibitem{ZhuL0W0W24}
L.~Zhu, B.~Liao, Q.~Zhang, X.~Wang, W.~Liu, and X.~Wang, ``Vision mamba: Efficient visual representation learning with bidirectional state space model,'' in \emph{Proc. Int. Conf. Mach. Learn. (ICML)}, 2024.

\bibitem{abs-2401-04722}
J.~Ma, F.~Li, and B.~Wang, ``U-mamba: Enhancing long-range dependency for biomedical image segmentation,'' \emph{arXiv preprint arXiv:2401.04722}, 2024.

\bibitem{abs-2412-20066}
B.~Li, H.~Zhao, W.~Wang, P.~Hu, Y.~Gou, and X.~Peng, ``Mair: {A} locality- and continuity-preserving mamba for image restoration,'' \emph{arXiv preprint arXiv:2412.20066}, 2024.

\bibitem{abs-2411-15269}
H.~Guo, Y.~Guo, Y.~Zha, Y.~Zhang, W.~Li, T.~Dai, S.~Xia, and Y.~Li, ``Mambairv2: Attentive state space restoration,'' \emph{arXiv preprint arXiv:2411.15269}, 2024.

\bibitem{MEAS}
X.~Yu, S.~Zhou, H.~Li, and L.~Zhu, ``Multi-expert adaptive selection: Task-balancing for all-in-one image restoration,'' \emph{arXiv preprint arXiv:2407.19139}, 2024.

\bibitem{LiCT19}
R.~Li, L.~Cheong, and R.~T. Tan, ``Heavy rain image restoration: Integrating physics model and conditional adversarial learning,'' in \emph{Proc. IEEE Conf. Comput. Vis. Pattern Recognit. (CVPR)}, 2019, pp. 1633--1642.

\bibitem{LiuJHH18}
Y.~Liu, D.~Jaw, S.~Huang, and J.~Hwang, ``Desnownet: Context-aware deep network for snow removal,'' \emph{{IEEE} Trans. Image Process.}, vol.~27, no.~6, pp. 3064--3073, 2018.

\bibitem{QianTYS018}
R.~Qian, R.~T. Tan, W.~Yang, J.~Su, and J.~Liu, ``Attentive generative adversarial network for raindrop removal from a single image,'' in \emph{Proc. IEEE Conf. Comput. Vis. Pattern Recognit. (CVPR)}, 2018, pp. 2482--2491.

\bibitem{ZamirA0HK0021}
S.~W. Zamir, A.~Arora, S.~H. Khan, M.~Hayat, F.~S. Khan, M.~Yang, and L.~Shao, ``Multi-stage progressive image restoration,'' in \emph{Proc. IEEE Conf. Comput. Vis. Pattern Recognit. (CVPR)}, 2021, pp. 14\,821--14\,831.

\bibitem{ChenCZS22}
L.~Chen, X.~Chu, X.~Zhang, and J.~Sun, ``Simple baselines for image restoration,'' in \emph{Proc. Eur. Conf. Comput. Vis. (ECCV)}, vol. 13667, 2022, pp. 17--33.

\bibitem{WangCBZLL22}
Z.~Wang, X.~Cun, J.~Bao, W.~Zhou, J.~Liu, and H.~Li, ``Uformer: {A} general u-shaped transformer for image restoration,'' in \emph{Proc. IEEE/CVF Conf. Comput. Vis. Pattern Recognit. (CVPR)}, 2022, pp. 17\,662--17\,672.

\bibitem{LiFXDRTG23}
Y.~Li, Y.~Fan, X.~Xiang, D.~Demandolx, R.~Ranjan, R.~Timofte, and L.~V. Gool, ``Efficient and explicit modelling of image hierarchies for image restoration,'' in \emph{Proc. IEEE/CVF Conf. Comput. Vis. Pattern Recognit. (CVPR)}, 2023, pp. 18\,278--18\,289.

\bibitem{ValanarasuYP22}
J.~M.~J. Valanarasu, R.~Yasarla, and V.~M. Patel, ``Transweather: Transformer-based restoration of images degraded by adverse weather conditions,'' in \emph{Proc. IEEE Conf. Comput. Vis. Pattern Recognit. (CVPR)}, 2022, pp. 2353--2363.

\bibitem{OzdenizciL23}
O.~{\"{O}}zdenizci and R.~Legenstein, ``Restoring vision in adverse weather conditions with patch-based denoising diffusion models,'' \emph{{IEEE} Trans. Pattern Anal. Mach. Intell.}, vol.~45, no.~8, pp. 10\,346--10\,357, 2023.

\bibitem{abs-2411-16739}
J.~Guo, H.~Yang, M.~Zhou, and X.~Zhang, ``Gradient-guided parameter mask for multi-scenario image restoration under adverse weather,'' \emph{arXiv preprint arXiv:2411.16739}, 2024.

\bibitem{ZhengWYZH024}
D.~Zheng, X.~Wu, S.~Yang, J.~Zhang, J.~Hu, and W.~Zheng, ``Selective hourglass mapping for universal image restoration based on diffusion model,'' in \emph{Proc. IEEE/CVF Conf. Comput. Vis. Pattern Recognit. (CVPR)}, 2024, pp. 25\,445--25\,455.

\bibitem{zhang2024efficient}
R.~Zhang, Y.~Luo, J.~Liu, H.~Yang, Z.~Dong, D.~Gudovskiy, T.~Okuno, Y.~Nakata, K.~Keutzer, Y.~Du \emph{et~al.}, ``Efficient deweahter mixture-of-experts with uncertainty-aware feature-wise linear modulation,'' in \emph{Proc. AAAI Conf. Artif. Intell.}, vol.~38, no.~15, 2024, pp. 16\,812--16\,820.

\bibitem{wang2024gridformer}
G.~R. dense transformer with grid structure for image restoration in adverse~weather conditions, ``Gridformer: Residual dense transformer with grid structure for image restoration in adverse weather conditions,'' \emph{Int. J. Comput. Vis.}, pp. 1--23, 2024.

\bibitem{Quan0L021}
R.~Quan, X.~Yu, Y.~Liang, and Y.~Yang, ``Removing raindrops and rain streaks in one go,'' in \emph{Proc. IEEE Conf. Comput. Vis. Pattern Recognit. (CVPR)}, 2021, pp. 9147--9156.

\end{thebibliography}

\vfill
\end{CJK}
\end{document}